\documentclass{article} 
\usepackage{iclr2021_conference,times}

\usepackage{amsmath,amsfonts,bm}

\def\1{\bm{1}}

\DeclareMathAlphabet{\mathsfit}{\encodingdefault}{\sfdefault}{m}{sl}
\SetMathAlphabet{\mathsfit}{bold}{\encodingdefault}{\sfdefault}{bx}{n}

\usepackage[
colorlinks=true,
breaklinks=true,
urlcolor=magenta,
linkcolor=red,
citecolor=blue,
]{hyperref}
\usepackage{url}

\usepackage{booktabs}
\usepackage{nicefrac}
\usepackage{microtype}
\usepackage{bbm}
\usepackage{amssymb}
\usepackage{graphicx}
\usepackage{subcaption}
\usepackage{multirow}
\usepackage{makecell}
\usepackage{diagbox}
\usepackage{algorithm}
\usepackage{algorithmic}
\usepackage{fancyvrb}
\usepackage[frozencache,cachedir=.]{minted} 
\usepackage{pifont}
\usepackage{tablefootnote}

\newcommand{\cmark}{\ding{51}}

\newcommand{\imix}{{$i$-Mix}}
\newcommand{\imixtext}{{i-Mix}}
\newcommand{\imixup}{{$i$-MixUp}}
\newcommand{\icutmix}{{$i$-CutMix}}
\newcommand{\inputmix}{InputMix}
\newcommand{\selfsl}{SSL}

\def\barleftwidth{0.436\linewidth}
\def\barrightwidth{0.545\linewidth}
\def\scatterleftwidth{0.436\linewidth}
\def\scatterrightwidth{0.545\linewidth}

\def\tftablewidth{0.436\linewidth}
\def\tffigwidth{0.545\linewidth}

\usepackage{titlesec} 
\titlespacing*{\paragraph} {0pt}{2pt}{4pt}

\iftrue 
    \newcommand{\cutabstractup}{\vspace{0pt}}
    \newcommand{\cutabstractdown}{\vspace{0pt}}

    \newcommand{\cutfigurecaptionup}{\vspace{0pt}}
    \newcommand{\cutfigurecaptiondown}{\vspace{0pt}}

    \newcommand{\cuttablecaptionup}{\vspace{0pt}}
    \newcommand{\cuttablecaptiondown}{\vspace{0pt}}
\else 
    \newcommand{\cutabstractup}{}
    \newcommand{\cutabstractdown}{}

    \newcommand{\cutfigurecaptionup}{}
    \newcommand{\cutfigurecaptiondown}{}

    \newcommand{\cuttablecaptionup}{}
    \newcommand{\cuttablecaptiondown}{}
\fi

\title{{\imix}: A Domain-Agnostic Strategy\\for Contrastive Representation Learning}

\author{%
  Kibok Lee\footnotemark[1]\,\,\textsuperscript{,}\footnotemark[2]\quad
  Yian Zhu\footnotemark[1]\quad
  Kihyuk Sohn\footnotemark[3]\quad
  Chun-Liang Li\footnotemark[3]\quad
  Jinwoo Shin\footnotemark[4]\quad
  Honglak Lee\footnotemark[1]\,\,\textsuperscript{,}\footnotemark[5]\\
  \footnotemark[1]\enspace University of Michigan\,
  \footnotemark[2]\enspace Amazon Web Services\,
  \footnotemark[3]\enspace Google Cloud AI\,
  \footnotemark[4]\enspace KAIST\,
  \footnotemark[5]\enspace LG AI Research\\
  \footnotemark[1]\enspace \texttt{\{kibok,yianz,honglak\}@umich.edu}\quad
  \footnotemark[3]\enspace \texttt{\{kihyuks,chunliang\}@google.com}\\
  \footnotemark[2]\enspace \texttt{kibok@amazon.com}\quad
  \footnotemark[4]\enspace \texttt{jinwoos@kaist.ac.kr}\quad
  \footnotemark[5]\enspace \texttt{honglak@lgresearch.ai}
}

\iclrfinalcopy 
\begin{document}

\maketitle

\begin{abstract}
\cutabstractup
Contrastive representation learning has shown to be effective to learn representations from unlabeled data. However, much progress has been made in vision domains relying on data augmentations carefully designed using domain knowledge. In this work, we propose {\imix}, a simple yet effective domain-agnostic regularization strategy for improving contrastive representation learning. We cast contrastive learning as training a non-parametric classifier by assigning a unique virtual class to each data in a batch. Then, data instances are mixed in both the input and virtual label spaces, providing more augmented data during training. In experiments, we demonstrate that {\imix} consistently improves the quality of learned representations across domains, including image, speech, and tabular data. Furthermore, we confirm its regularization effect via extensive ablation studies across model and dataset sizes. The code is available at \textcolor{magenta}{\url{https://github.com/kibok90/imix}}.
\cutabstractdown
\end{abstract}

\section{Introduction}
\label{sec:intro}

Representation learning~\citep{bengio2013representation} is a fundamental task in machine learning since the success of machine learning relies on the quality of representation.
Self-supervised representation learning ({\selfsl}) has been successfully applied in several domains, including
image recognition~\citep{he2020momentum,chen2020simple},
%
natural language processing~\citep{mikolov2013distributed,devlin2018bert},
robotics~\citep{sermanet2018time,lee2019making},
%
speech recognition~\citep{ravanelli2020multi}, and
video understanding~\citep{korbar2018cooperative,owens2018audio}.
%
Since no label is available in the unsupervised setting, pretext tasks are proposed to provide self-supervision:
for example,
%
%
%
%
context prediction~\citep{doersch2015unsupervised},
inpainting~\citep{pathak2016context}, and
contrastive learning~\citep{wu2018unsupervised,hjelm2019learning,he2020momentum,chen2020simple}.
%
{\selfsl} has also been used as an auxiliary task to improve the performance on the main task, such as
generative model learning~\citep{chen2019self},
semi-supervised learning~\citep{zhai2019s4l}, and
improving robustness and uncertainty~\citep{hendrycks2019using}.

Recently, contrastive representation learning has gained increasing attention by showing state-of-the-art performance in {\selfsl} for large-scale image recognition~\citep{he2020momentum,chen2020simple}, which outperforms its supervised pre-training counterpart~\citep{he2016deep} on downstream tasks.
However, while the concept of contrastive learning is applicable to any domains, the quality of learned representations rely on the domain-specific inductive bias:
as anchors and positive samples are obtained from the same data instance, data augmentation introduces semantically meaningful variance for better generalization.
To achieve a strong, yet semantically meaningful data augmentation, domain knowledge is required, e.g., color jittering in 2D images or structural information in video understanding.
Hence, contrastive representation learning in different domains requires an effort to develop effective data augmentations.
Furthermore, while recent works have focused on large-scale settings where millions of unlabeled data is available, it would not be practical in real-world applications.
For example, in lithography, acquiring data is very expensive in terms of both time and cost due to the complexity of manufacturing process~\citep{lin2018data,sim2019automatic}.

Meanwhile, MixUp~\citep{zhang2018mixup} has shown to be a successful data augmentation for supervised learning in various domains and tasks, including
image classification~\citep{zhang2018mixup},
generative model learning~\citep{lucas2018mixed}, and
natural language processing~\citep{guo2019augmenting,guo2020nonlinear}.
In this paper, we explore the following natural, yet important question:
is the idea of MixUp useful for unsupervised, self-supervised, or contrastive representation learning across different domains?

To this end,  
we propose \emph{instance Mix ({\imix})},
a domain-agnostic regularization strategy for contrastive representation learning.
The key idea of {\imix} is to introduce virtual labels in a batch and mix data instances and their corresponding virtual labels in the input and label spaces, respectively.
We first introduce the general formulation of {\imix}, and then we show the applicability of {\imix} to state-of-the-art contrastive representation learning methods, SimCLR~\citep{chen2020simple} and MoCo~\citep{he2020momentum}, and a self-supervised learning method
without negative pairs, BYOL~\citep{grill2020bootstrap}.

Through the experiments, we demonstrate the efficacy of {\imix} in a variety of settings.
First, we show the effectiveness of {\imix} by evaluating the discriminative performance of learned representations in multiple domains.
Specifically, we adapt {\imix} to the contrastive representation learning methods,
advancing state-of-the-art performance across different domains, including image~\citep{krizhevsky2009learning,deng2009imagenet}, speech~\citep{warden2018speech}, and tabular~\citep{asuncion2007uci} datasets.
Then, we study {\imix} in various conditions, including when
1)~the model and training dataset is small or large,
2)~domain knowledge is limited, and 
3)~transfer learning.


\paragraph{Contribution.} In summary, our contribution is three-fold:
\begin{itemize}
\item We propose {\imix}, a method for regularizing contrastive representation learning, motivated by MixUp~\citep{zhang2018mixup}.
%
We show how to apply {\imix} to state-of-the-art contrastive representation learning methods~\citep{chen2020simple,he2020momentum,grill2020bootstrap}.
%
%
%
\item We show that {\imix} consistently improves contrastive representation learning in both vision and non-vision domains.
In particular, the discriminative performance of representations learned with {\imix} is on par with fully supervised learning on CIFAR-10/100~\citep{krizhevsky2009learning} and Speech Commands~\citep{warden2018speech}.
\item We verify the regularization effect of {\imix} in a variety of settings.
We empirically observed that {\imix} significantly improves contrastive representation learning when
1) the training dataset size is small, or
2) the domain knowledge for data augmentations is not enough.
%
%
\end{itemize}

\section{Related Work}
\label{sec:related}

\paragraph{Self-supervised representation learning ({\selfsl})}
aims at learning representations from unlabeled data by solving a pretext task that is derived from self-supervision.
%
Early works on {\selfsl} proposed pretext tasks based on data reconstruction by autoencoding~\citep{bengio2007greedy}, such as 
%
%
context prediction~\citep{doersch2015unsupervised} and
inpainting~\citep{pathak2016context}.
Decoder-free {\selfsl} has made a huge progress in recent years.
Exemplar CNN~\citep{dosovitskiy2014discriminative} learns by classifying individual instances with data augmentations.
%
%
{\selfsl} of visual representation, including
colorization~\citep{zhang2016colorful},
solving jigsaw puzzles~\citep{noroozi2016unsupervised},
counting the number of objects~\citep{noroozi2017representation},
rotation prediction~\citep{gidaris2018unsupervised},
next pixel prediction~\citep{oord2018representation,henaff2019data}, and
combinations of them~\citep{doersch2017multi,kim2018learning,noroozi2018boosting} often leverages image-specific properties to design pretext tasks.
Meanwhile, alhough deep clustering~\citep{caron2018deep,caron2019unsupervised,asano2020self} is often distinguished from {\selfsl},
it also leverages unsupervised clustering assignments as self-supervision for representation learning.

%

\paragraph{Contrastive representation learning}
has gained lots of attention for
{\selfsl}~\citep{he2020momentum,chen2020simple}.
As opposed to early works on exemplar CNN~\citep{dosovitskiy2014discriminative,dosovitskiy2015discriminative},
contrastive learning
maximizes similarities of positive pairs while minimizes similarities of negative pairs
instead of training an instance classifier.
As the choice of negative pairs is crucial for the quality of learned representations, recent works have carefully designed them.
Memory-based approaches~\citep{wu2018unsupervised,hjelm2019learning,bachman2019learning,misra2020self,tian2019contrastive} maintain a memory bank of embedding vectors of instances to keep negative samples, where the memory is updated with embedding vectors extracted from previous batches.
In addition, MoCo~\citep{he2020momentum}
%
showed that differentiating the model for anchors and positive/negative samples is effective, where the model for positive/negative samples is updated by the exponential moving average of the model for anchors.
On the other hand, recent works~\citep{ye2019unsupervised,misra2020self,chen2020simple,tian2019contrastive} showed that learning invariance to different views is important in contrastive representation learning.
The views can be generated through data augmentations carefully designed using domain knowledge~\citep{chen2020simple}, splitting input channels~\citep{tian2019contrastive}, or borrowing the idea of other pretext tasks, such as creating jigsaw puzzles or rotating inputs~\citep{misra2020self}.
In particular, SimCLR~\citep{chen2020simple} showed that a simple memory-free approach with a large batch size and strong data augmentations has a comparable performance to memory-based approaches.
InfoMin~\citep{tian2020makes} further studied a way to generate good views for contrastive representation learning and achieved state-of-the-art performance by combining prior works.
Different from other contrastive representation learning methods, BYOL~\citep{grill2020bootstrap} does not require negative pairs, where the proposed pretext task aims at predicting latent representations of one view from another.
%
While prior works have focused on {\selfsl} on large-scale visual recognition tasks, our work focuses on contrastive representation learning in both small- and large-scale settings in different domains.





\paragraph{Data augmentation}
%
is a technique to increase the diversity of data, especially when training data are not enough for generalization.
Since the augmented data must be understood as the original data, data augmentations are carefully designed using the domain knowledge about images~\citep{devries2017improved,cubuk2019autoaugment,cubuk2019randaugment,zhong2020random}, speech~\citep{amodei2016deep,park2019specaugment}, or natural languages~\citep{zhang2015character,wei2019eda}.

Some works have studied data augmentation with less domain knowledge:
\citet{devries2017dataset} proposed a domain-agnostic augmentation strategy by first encoding the dataset and then applying augmentations in the feature space.
MixUp~\citep{zhang2018mixup} is an effective data augmentation strategy in supervised learning,
which performs vicinal risk minimization instead of empirical risk minimization, by linearly interpolating input data and their labels on the data and label spaces, respectively.
On the other hand, MixUp has also shown its effectiveness in other tasks and non-vision domains, including
generative adversarial networks~\citep{lucas2018mixed},
improved robustness and uncertainty~\citep{hendrycks2020augmix}, and
sentence classification in natural language processing~\citep{guo2020nonlinear,guo2019augmenting}.
Other variations have also been investigated by interpolating in the feature space~\citep{verma2019manifold} or leveraging domain knowledge~\citep{yun2019cutmix}.
MixUp would not be directly applicable to some domains, such as point clouds, 
but its adaptation can be effective~\citep{harris2020fmix}.
{\imix} is a kind of data augmentation for better generalization in contrastive representation learning, resulting in better performances on downstream tasks.

\paragraph{Concurrent works}
have leveraged the idea of MixUp for contrastive representation learning.
As discussed in Section~\ref{sec:inputmix}, only input data can be mixed for improving contrastive representation learning~\citep{shen2020rethinking,verma2020towards,zhou2020comparing}, which can be considered as injecting data-driven noises.
%
\citet{kalantidis2020hard} mixed hard negative samples on the embedding space. 
\citet{kim2020mixco} reported similar observations to ours but focused on small image datasets.

\section{Approach}
\label{sec:approach}

In this section, we review MixUp~\citep{zhang2018mixup} in supervised learning and present {\imix} in contrastive learning~\citep{he2020momentum,chen2020simple,grill2020bootstrap}.
Throughout this section, let $\mathcal{X}$ be a data space, $\mathbb{R}^{D}$ be a $D$-dimensional embedding space, and a model $f \,{:}\, \mathcal{X} \,{\rightarrow}\, \mathbb{R}^{D}$ be a mapping between them.
For conciseness, $f_{i} \,{=}\, f(x_{i})$ and $\tilde{f}_{i} \,{=}\, f(\tilde{x}_{i})$ for $x_{i}, \tilde{x}_{i} \in \mathcal{X}$, and model parameters are omitted in loss functions.

\subsection{MixUp in Supervised Learning}
\label{sec:mixup}

Suppose an one-hot label $y_{i} \in \{0,1\}^{C}$ is assigned to a data $x_{i}$, where $C$ is the number of classes.
%
Let a linear classifier predicting the labels consists of weight vectors $\{w_{1}, \dots, w_{C}\}$, where $w_{c} \in \mathbb{R}^{D}$.\footnote{We omit bias terms for presentation clarity.}
Then, the cross-entropy loss for supervised learning is defined as:
\begin{align}
\ell_{\text{Sup}}(x_{i}, y_{i}) =
-\sum_{c=1}^{C} y_{i,c}
\log \frac{ \exp(w_{c}^{\top} f_{i}) }{ \sum_{k=1}^{C} \exp( w_{k}^{\top} f_{i}) }.
\label{eq:sup}
\end{align}
While the cross-entropy loss is widely used for supervised training of deep neural networks, there are several challenges of training with the cross-entropy loss, such as preventing overfitting or networks being overconfident.
Several regularization techniques have been proposed to alleviate these issues, including
label smoothing~\citep{szegedy2016rethinking},
%
%
adversarial training~\citep{miyato2018virtual}, and
confidence calibration~\citep{lee2018training}.
%
%

MixUp~\citep{zhang2018mixup} is an effective regularization with negligible computational overhead.
It conducts a linear interpolation of two data instances in both input and label spaces and trains a model by minimizing the cross-entropy loss defined on the interpolated data and labels.
Specifically, for two labeled data $(x_{i}, y_{i})$, $(x_{j}, y_{j})$, the MixUp loss is defined as follows:
\begin{align}
\ell_{\text{Sup}}^{\text{MixUp}} \big( (x_{i}, y_{i}), (x_{j}, y_{j}) ; \lambda \big) =
\ell_{\text{Sup}}( \lambda x_{i} + (1 - \lambda) x_{j}, \lambda y_{i} + (1 - \lambda) y_{j} ),
\label{eq:mixup}
\end{align}
where $\lambda \,{\sim}\, \text{Beta}(\alpha, \alpha)$ is a mixing coefficient sampled from the beta distribution.
MixUp is a vicinal risk minimization method~\citep{chapelle2001vicinal} that augments data and their labels
in a data-driven manner.
Not only improving the generalization on the supervised task, it also improves adversarial robustness~\citep{pang2019mixup} and confidence calibration~\citep{thulasidasan2019mixup}.

\subsection{\texorpdfstring{\imix}{\imixtext} in Contrastive Learning}
\label{sec:imix}

We introduce \emph{instance mix ({\imix})}, a data-driven augmentation strategy for contrastive representation learning to improve the generalization of learned representations.
%
Intuitively, instead of mixing class labels, {\imix} interpolates their \emph{virtual} labels, which indicates their identity in a batch.

Let $\mathcal{B} \,{=}\, \{(x_{i}, \tilde{x}_{i})\}_{i=1}^{N}$ be a batch of data pairs, where $N$ is the batch size, $x_{i}, \tilde{x}_{i} \,{\in}\, \mathcal{X}$ are two views of the same data, which are usually generated by different augmentations.
For each anchor $x_{i}$, we call $\tilde{x}_{i}$ and $\tilde{x}_{j {\neq} i}$ positive and negative samples, respectively.\footnote{%
Some literature~\citep{he2020momentum,chen2020improved} refers to them as query and positive/negative keys.
}
Then, the model $f$ learns to maximize similarities of positive pairs (instances from the same data) while minimize similarities of negative pairs (instances from different data) in the embedding space.
The output of $f$ is L2-normalized,
%
which has shown to be effective~\citep{wu2018incremental,he2020momentum,chen2020simple}.
Let $v_{i} \,{\in}\, \{0,1\}^{N}$ be the virtual label of $x_{i}$ and $\tilde{x}_{i}$ in a batch $\mathcal{B}$, where $v_{i,i} \,{=}\, 1$ and $v_{i,j {\neq} i} \,{=}\, 0$.
For a general sample-wise contrastive loss with virtual labels $\ell(x_{i}, v_{i})$, the {\imix} loss is defined as follows:
\begin{align}
\ell^{\text{\imix}} \big( (x_{i}, v_{i}), (x_{j}, v_{j}) ; \mathcal{B}, \lambda \big) =
\ell( \text{Mix}(x_{i}, x_{j}; \lambda), \lambda v_{i} + (1 - \lambda) v_{j} ; \mathcal{B}),
\label{eq:imix}
\end{align}
where $\lambda \,{\sim}\, \text{Beta}(\alpha, \alpha)$ is a mixing coefficient and $\text{Mix}$ is a mixing operator, which can be adapted depending on target domains:
for example,
$\text{MixUp}(x_{i}, x_{j}; \lambda) \,{=}\, \lambda x_{i} \,{+}\, (1 {-} \lambda) x_{j}$~\citep{zhang2018mixup} when data values are continuous,
and $\text{CutMix}(x_{i}, x_{j}; \lambda) \,{=}\, M_{\lambda} {\odot} x_{i} \,{+}\, (1{-}M_{\lambda}) {\odot} x_{j}$~\citep{yun2019cutmix} when data values have a spatial correlation with neighbors, where $M_{\lambda}$ is a binary mask filtering out a region whose relative area is $(1{-}\lambda)$, and $\odot$ is an element-wise multiplication.
Note that some mixing operators might not work well for some domains:
%
for example, CutMix would not be valid when data values and their spatial neighbors have no correlation.
However, the MixUp operator generally works well across domains including image, speech, and tabular;
we use it for {\imix} formulations and experiments, unless otherwise specified.
%
%
In the following, we show how to apply {\imix} to contrastive representation learning methods.

\paragraph{SimCLR~\citep{chen2020simple}}
is a simple contrastive representation learning method without a memory bank, where
%
each anchor has one positive sample and ($2N{-}2$) negative samples.
Let $x_{N+i} = \tilde{x}_{i}$ for conciseness.
Then, the ($2N{-}1$)-way discrimination loss is written as follows:
\begin{align}
\ell_{\text{SimCLR}} ( x_{i} ; \mathcal{B} ) =
-\log \frac{ \exp \big( s(f_{i}, f_{(N+i) \bmod 2N}) / \tau \big) }
{ \sum_{k=1, k \neq i}^{2N} \exp \big( s(f_{i}, f_{k}) / \tau \big) },
\label{eq:simclr}
\end{align}
where $\tau$ is a temperature scaling parameter and $s(f, \tilde{f}) \,{=}\, (f^{\top} \tilde{f}) / \lVert f \rVert \lVert \tilde{f} \rVert$ is the inner product of two L2-normalized vectors.
%
In this formulation, {\imix} is not directly applicable because virtual labels are defined differently for each anchor.\footnote{%
We present the application of {\imix} to the original SimCLR formulation in Appendix~\ref{sec:imix_more}.
}
To resolve this issue,
%
we simplify the formulation of SimCLR by excluding anchors from negative samples.
Then, with virtual labels, the $N$-way discrimination loss is written as follows:
\begin{align}
\ell_{\text{N-pair}} ( x_{i}, v_{i} ; \mathcal{B} ) =
-\sum_{n=1}^{N} v_{i,n}
\log \frac{ \exp \big( s(f_{i}, \tilde{f}_{n}) / \tau \big) }
{ \sum_{k=1}^{N} \exp \big( s(f_{i}, \tilde{f}_{k}) / \tau \big) },
\label{eq:npair}
\end{align}
where we call it the \textbf{N-pair} contrastive loss, as the formulation is similar to the N-pair loss in the context of metric learning~\citep{sohn2016improved}.%
\footnote{%
InfoNCE~\citep{oord2018representation} is a similar loss inspired by the idea of noise-contrastive estimation~\citep{gutmann2010noise}.
}
For two data instances $(x_{i}, v_{i})$, $(x_{j}, v_{j})$ and a batch of data pairs $\mathcal{B} \,{=}\, \{(x_{i}, \tilde{x}_{i})\}_{i=1}^{N}$, the {\imix} loss is defined as follows:
\begin{align}
\ell_{\text{N-pair}}^{\text{\imix}} \big( (x_{i}, v_{i}), (x_{j}, v_{j}) ; \mathcal{B}, \lambda \big) =
\ell_{\text{N-pair}}( \lambda x_{i} + (1 - \lambda) x_{j}, \lambda v_{i} + (1 - \lambda) v_{j} ; \mathcal{B}).
\label{eq:npair_imix}
\end{align}


\begin{algorithm}[t]
\caption{Loss computation for {\imix} on N-pair contrastive learning in PyTorch-like style.}
\label{alg:imix}
\begin{minted}[fontfamily=courier]{python}
a, b = aug(x), aug(x) # two different views of input x
lam = Beta(alpha, alpha).sample() # mixing coefficient
randidx = randperm(len(x))
a = lam * a + (1-lam) * a[randidx]
logits = matmul(normalize(model(a)), normalize(model(b)).T) / t
loss = lam * CrossEntropyLoss(logits, arange(len(x))) + \
       (1-lam) * CrossEntropyLoss(logits, randidx)
\end{minted}
\end{algorithm}

Algorithm~\ref{alg:imix} provides the pseudocode of {\imix} on N-pair contrastive learning for one iteration.\footnote{%
For losses linear with respect to labels (e.g., the cross-entropy loss), they are equivalent to
$\lambda \ell( \lambda x_{i} + (1 - \lambda) x_{j},  v_{i} ) + (1 - \lambda) \ell( \lambda x_{i} + (1 - \lambda) x_{j}, v_{j} )$, i.e., optimization to the mixed label is equivalent to joint optimization to original labels.
The proof for losses in contrastive learning methods is provided in Appendix~\ref{sec:linear}.
}

\paragraph{Pair relations in contrastive loss.}
To use contrastive loss for representation learning, one needs to properly define a pair relation $\{(x_{i}, \tilde{x}_{i})\}_{i=1}^{N}$.
For contrastive representation learning,
where semantic class labels are not provided, the pair relation would be defined in that
1) a positive pair, $x_{i}$ and $\tilde{x}_{i}$, are different views of the same data and
2) a negative pair, $x_{i}$ and $\tilde{x}_{j \neq i}$, are different data instances.
For supervised representation learning,
$x_{i}$ and $\tilde{x}_{i}$ are two data instances from the same class, while $x_{i}$ and $\tilde{x}_{j \neq i}$ are from different classes.
Note that two augmented versions of the same data also belong to the same class,
so they can also be considered as a positive pair.
{\imix} is not limited to self-supervised contrastive representation learning, but it can also be used as a regularization method for supervised contrastive representation learning~\citep{khosla2020supervised} or deep metric learning~\citep{sohn2016improved,movshovitz2017no}.

\paragraph{MoCo~\citep{he2020momentum}.}
%
%
In contrastive representation learning, the number of negative samples affects the quality of learned representations~\citep{arora2019theoretical}.
Because SimCLR mines negative samples in the current batch, having a large batch size is crucial, which often requires a lot of computational resources~\citep{chen2020simple}.
For efficient training, recent works have maintained a memory bank $\mathcal{M} \,{=}\, \{\mu_{k}\}_{k=1}^{K}$, which is a queue of previously extracted embedding vectors, where $K$ is the size of the memory bank~\citep{wu2018unsupervised,he2020momentum,tian2019contrastive,tian2020makes}.
In addition, MoCo introduces an exponential moving average (EMA) model to extract positive and negative embedding vectors, whose parameters are updated as
%
$\theta_{f^{\texttt{EMA}}} \,{\leftarrow}\, m \theta_{f^{\texttt{EMA}}} \,{+}\, (1 \,{-}\, m) \theta_{f}$, where $m \,{\in}\, [0,1)$ is a momentum coefficient and $\theta$ is model parameters.
%
The loss is written as follows:
%
\begin{align}
\ell_{\text{MoCo}} ( x_{i} ; \mathcal{B}, \mathcal{M} ) =
-\log \frac{ \exp \big( s(f_{i}, \tilde{f}_{i}^{\texttt{EMA}}) / \tau \big) }
{ \exp \big( s(f_{i}, \tilde{f}_{i}^{\texttt{EMA}}) / \tau \big)
+ \sum_{k=1}^{K} \exp \big( s(f_{i}, \mu_{k}) / \tau \big) }.
\label{eq:moco}
\end{align}
The memory bank $\mathcal{M}$ is then updated with $\{\tilde{f}_{i}^{\texttt{EMA}}\}$ in the first-in first-out order.
In this ($K{+}1$)-way discrimination loss, data pairs are independent to each other,
such that {\imix} is not directly applicable because virtual labels are defined differently for each anchor.
To overcome this issue, we include the positive samples of other anchors as negative samples, similar to the N-pair contrastive loss in Eq.~\eqref{eq:npair}.
%
Let $\tilde{v}_{i} \,{\in}\, \{0,1\}^{N+K}$ be a virtual label indicating the positive sample of each anchor, where $\tilde{v}_{i,i} \,{=}\, 1$ and $\tilde{v}_{i,j \neq i} \,{=}\, 0$.
%
%
Then, the ($N{+}K$)-way discrimination loss is written as follows:
%
\begin{align}
\ell_{\text{MoCo}} ( x_{i}, \tilde{v}_{i} ; \mathcal{B}, \mathcal{M} ) =
-\sum_{n=1}^{N} \tilde{v}_{i,n}
\log \frac{ \exp \big( s(f_{i}, \tilde{f}_{n}^{\texttt{EMA}}) / \tau \big) }
{ \sum_{k=1}^{N} \exp \big( s(f_{i}, \tilde{f}_{k}^{\texttt{EMA}}) / \tau \big)
+ \sum_{k=1}^{K} \exp \big( s(f_{i}, \mu_{k}) / \tau \big) }.
\label{eq:moco_v}
\end{align}
%
%
As virtual labels are bounded in the same set in this formulation, {\imix} is directly applicable:
for two data instances $(x_{i}, \tilde{v}_{i})$, $(x_{j}, \tilde{v}_{j})$, a batch of data pairs $\mathcal{B} \,{=}\, \{(x_{i}, \tilde{x}_{i})\}_{i=1}^{N}$, and the memory bank $\mathcal{M}$, the {\imix} loss is defined as follows:
\begin{align}
\ell_{\text{MoCo}}^{\text{\imix}} \big( (x_{i}, \tilde{v}_{i}), (x_{j}, \tilde{v}_{j}) ; \mathcal{B}, \mathcal{M}, \lambda \big) = 
\ell_{\text{MoCo}}( \lambda x_{i} + (1 - \lambda) x_{j}, \lambda \tilde{v}_{i} + (1 - \lambda) \tilde{v}_{j} ; \mathcal{B}, \mathcal{M} ).
\label{eq:moco_imix}
\end{align}

\paragraph{BYOL~\citep{grill2020bootstrap}.}
Different from other contrastive representation learning methods, BYOL is a self-supervised representation learning method without contrasting negative pairs.
For two views of the same data $x_{i}, \tilde{x}_{i} \,{\in}\, \mathcal{X}$, the model $f$ learns to predict a view embedded with the EMA model $\tilde{f}_{i}^{\texttt{EMA}}$ from its embedding $f_{i}$.
Specifically, an additional prediction layer $g$ is introduced, such that the difference between $g(f_{i})$ and $\tilde{f}_{i}^{\texttt{EMA}}$ is learned to be minimized.
%
%
The BYOL loss is written as follows:
%
\begin{align}
\ell_{\text{BYOL}} ( x_{i}, \tilde{x}_{i} ) =
\left\lVert g(f_{i}) / \lVert g(f_{i}) \rVert - \tilde{f}_{i} / \lVert \tilde{f}_{i} \rVert \right\rVert^{2}
= 2 - 2 \cdot s(g(f_{i}), \tilde{f}_{i}).
\label{eq:byol}
\end{align}
This formulation can be represented in the form of the general contrastive loss in Eq.~\eqref{eq:imix}, as the second view $\tilde{x}_{i}$ can be accessed from the batch $\mathcal{B}$ with its virtual label $v_{i}$.
To derive {\imix} in BYOL, let $\tilde{F} \,{=}\, [\tilde{f}_{1} / \lVert \tilde{f}_{1} \rVert, {\dots}, \tilde{f}_{N} / \lVert \tilde{f}_{N} \rVert] \,{\in}\, \mathbb{R}^{D {\times} N}$ be the collection of L2-normalized embedding vectors of the second views, such that $\tilde{f}_{i} / \lVert \tilde{f}_{i} \rVert \,{=}\, \tilde{F} v_{i}$.
Then, the BYOL loss is written as follows:
%
\begin{align}
\ell_{\text{BYOL}} ( x_{i}, v_{i} ; \mathcal{B} ) = 
\left\lVert g(f_{i}) / \lVert g(f_{i}) \rVert - \tilde{F} v_{i} \right\rVert^{2}
\label{eq:byol_v}
= 2 - 2 \cdot s(g(f_{i}), \tilde{F} v_{i}).
\end{align}
For two data instances $(x_{i}, v_{i})$, $(x_{j}, v_{j})$ and a batch of data pairs $\mathcal{B} \,{=}\, \{(x_{i}, \tilde{x}_{i})\}_{i=1}^{N}$, the {\imix} loss is defined as follows:
\begin{align}
\ell_{\text{BYOL}}^{\text{\imix}} \big( (x_{i}, v_{i}), (x_{j}, v_{j}) ; \mathcal{B}, \lambda \big) =
\ell_{\text{BYOL}}( \lambda x_{i} + (1 - \lambda) x_{j}, \lambda v_{i} + (1 - \lambda) v_{j} ; \mathcal{B}).
\label{eq:byol_imix}
\end{align}



\subsection{{\inputmix}}
\label{sec:inputmix}

%
The contribution of data augmentations to the quality of learned representations is crucial in contrastive representation learning.
%
%
For the case when the domain knowledge about efficient data augmentations is limited,
we propose to apply {\inputmix} together with {\imix}, which mixes input data but not their labels.
%
%
This method can be viewed as introducing structured noises driven by auxiliary data to the principal data with the largest mixing coefficient $\lambda$, and the label of the principal data is assigned to the mixed data~\citep{shen2020rethinking,verma2020towards,zhou2020comparing}.
%
%
We applied {\inputmix} and {\imix} together on image datasets in Table~\ref{tb:aug}.

\section{Experiments}
\label{sec:exp}

In this section, we demonstrate the effectiveness of {\imix}. 
In all experiments, we conduct contrastive representation learning on a pretext dataset and evaluate the quality of representations via supervised classification on a downstream dataset. We report the accuracy averaged over up to five runs.
In the first stage, a convolutional neural network (CNN) or multilayer perceptron (MLP) followed by the two-layer MLP projection head is trained on an unlabeled dataset.
Then, we replace the projection head with a linear classifier and train only the linear classifier on a labeled dataset for downstream task.
Except for transfer learning, datasets for the pretext and downstream tasks are the same.
%
For {\imix}, we sample a mixing coefficient $\lambda \,{\sim}\, \text{Beta}(\alpha,\alpha)$ for each data, where $\alpha \,{=}\, 1$ unless otherwise stated.%
\footnote{%
$\text{Beta}(\alpha,\alpha)$ is the uniform distribution when $\alpha \,{=}\, 1$, bell-shaped when $\alpha \,{>}\, 1$, and bimodal when $\alpha \,{<}\, 1$.
}
Additional details for the experimental settings and more experiments can be found in Appendix~\ref{sec:exp_more}.

\subsection{Experimental Setup}
\label{sec:setup}
\paragraph{Baselines and datasets.}
We consider
1) N-pair contrastive learning as a memory-free contrastive learning method,\footnote{%
We use the N-pair formulation in Eq.~\eqref{eq:npair} instead of SimCLR as it is simpler and more efficient to integrate {\imix}.
As shown in Appendix~\ref{sec:var}, the N-pair formulation results in no worse performance than SimCLR.
}
2) MoCo v2~\citep{he2020momentum,chen2020improved}
\footnote{%
MoCo v2 improves the performance of MoCo by cosine learning schedule and more data augmentations.
}
as a memory-based contrastive learning method, and
3) BYOL~\citep{grill2020bootstrap},
which is a self-supervised learning method without negative pairs.
We apply {\imix} to these methods and compare their performances.
To show the effectiveness of {\imix} across domains, we evaluate the methods on datasets from multiple domains, including 
image, speech, and tabular datasets.

%
CIFAR-10/100~\citep{krizhevsky2009learning} consist of 50k training and 10k test images, and ImageNet~\citep{deng2009imagenet} has 1.3M training and 50k validation images, where we use them for evaluation.
%
For ImageNet, we also use a subset of randomly chosen 100 classes out of 1k classes to experiment at a different scale.
We apply a set of data augmentations randomly in sequence including random resized cropping,
horizontal flipping, color jittering, gray scaling,
and Gaussian blurring for ImageNet,
which has shown to be effective~\citep{chen2020simple,chen2020improved}.
We use ResNet-50~\citep{he2016deep} as a backbone network.
Models are trained with a batch size of 256 (i.e., 512 including augmented data) 
%
for up to 4000 epochs on CIFAR-10 and 100,
and with a batch size of 512 for 800 epochs on ImageNet.
%
%
For ImageNet experiments, we use the CutMix~\citep{yun2019cutmix} version of {\imix}.

%
The Speech Commands dataset~\citep{warden2018speech} contains 51k training, 7k validation, and 7k test data
in 12 classes.
%
%
We apply a set of data augmentations randomly in sequence:
changing amplitude, speed, and pitch in time domain,
stretching, time shifting, and adding background noise in frequency domain.
Augmented data are then transformed to a $32{\times}32$ mel spectogram.
We use the same architecture with image experiments.
Models are trained with a batch size of 256 for 500 epochs.

%
For tabular dataset experiments, we consider Forest Cover Type (CovType) and Higgs Boson (Higgs) from UCI repository~\citep{asuncion2007uci}.
CovType contains 15k training and 566k test data in 7 classes, and
Higgs contains 10.5M training and 0.5M test data for binary classification.
For Higgs, we use a subset of 100k and 1M training data to experiment at a different scale.
Since the domain knowledge for data augmentations on tabular data is limited,
only a masking noise with the probability 0.2 is considered as a data augmentation.
We use a 5-layer MLP with batch normalization~\citep{ioffe2015batch} as a backbone network.
Models are trained with a batch size of 512 for 500 epochs.
We use $\alpha \,{=}\, 2$ for CovType and Higgs100k, as it is slightly better than $\alpha \,{=}\, 1$.

\subsection{Main Results}
\label{sec:domain}

\begin{table}[t]
\centering
\begin{tabular}{cccccccc}
\toprule
Domain & Dataset & N-pair & + {\imix} & MoCo v2 & + {\imix} & BYOL & + {\imix} \cr
\cmidrule(rl){1-1}\cmidrule(rl){2-2}\cmidrule(rl){3-4}\cmidrule(rl){5-6}\cmidrule(rl){7-8}
\multirow{2}{*}{Image} & CIFAR-10 &
{93.3} \scriptsize{$\pm$ 0.1} & \textbf{95.6} \scriptsize{$\pm$ 0.2} & {93.5} \scriptsize{$\pm$ 0.2} & \textbf{96.1} \scriptsize{$\pm$ 0.1} & {94.2} \scriptsize{$\pm$ 0.2} & \textbf{96.3} \scriptsize{$\pm$ 0.2} \cr
& CIFAR-100 &
{70.8} \scriptsize{$\pm$ 0.4} & \textbf{75.8} \scriptsize{$\pm$ 0.3} & {71.6} \scriptsize{$\pm$ 0.1} & \textbf{78.1} \scriptsize{$\pm$ 0.3} & {72.7} \scriptsize{$\pm$ 0.4} & \textbf{78.6} \scriptsize{$\pm$ 0.2} \cr
\cmidrule(rl){1-1}\cmidrule(rl){2-2}\cmidrule(rl){3-4}\cmidrule(rl){5-6}\cmidrule(rl){7-8}
Speech & Commands &
{94.9} \scriptsize{$\pm$ 0.1} & \textbf{98.3} \scriptsize{$\pm$ 0.1} & {96.3} \scriptsize{$\pm$ 0.1} & \textbf{98.4} \scriptsize{$\pm$ 0.0} & {94.8} \scriptsize{$\pm$ 0.2} & \textbf{98.3} \scriptsize{$\pm$ 0.0} \cr
\cmidrule(rl){1-1}\cmidrule(rl){2-2}\cmidrule(rl){3-4}\cmidrule(rl){5-6}\cmidrule(rl){7-8}
Tabular & CovType &
{68.5} \scriptsize{$\pm$ 0.3} & \textbf{72.1} \scriptsize{$\pm$ 0.2} & {70.5} \scriptsize{$\pm$ 0.2} & \textbf{73.1} \scriptsize{$\pm$ 0.1} & {72.1} \scriptsize{$\pm$ 0.2} & \textbf{74.1} \scriptsize{$\pm$ 0.2} \cr
\bottomrule
\end{tabular}
\cuttablecaptionup
\caption{
Comparison of contrastive representation learning methods and {\imix} in different domains.
%
}
\cuttablecaptiondown
\label{tb:domain}
\end{table}

Table~\ref{tb:domain} shows the wide applicability of {\imix} to state-of-the-art contrastive representation learning methods in multiple domains.
{\imix} results in consistent improvements on the classification accuracy, e.g., up to 6.5\% when {\imix} is applied to MoCo v2 on CIFAR-100.
Interestingly, we observe that linear classifiers on top of representations learned with {\imix} without fine-tuning the pre-trained part often yield
a classification accuracy on par with
simple end-to-end supervised learning from random initialization,
%
e.g., {\imix} vs. end-to-end supervised learning performance is 96.3\% vs. 95.5\% on CIFAR-10, 78.6\% vs. 78.9\% on CIFAR-100, and 98.2\% vs. 98.0\% on Speech Commands.%
\footnote{%
Supervised learning with improved methods, e.g., MixUp, outperforms $i$-Mix.
However, linear evaluation on top of self-supervised representation learning is a proxy to measure the quality of representations learned without labels, such that it is not supposed to be compared with the performance of supervised learning.
%
}


\subsection{Regularization Effect of \texorpdfstring{\imix}{\imixtext}}
\label{sec:scalablility}

\begin{figure}[t]
\centering
\begin{subfigure}[h]{\barleftwidth}
\includegraphics[width=\linewidth]{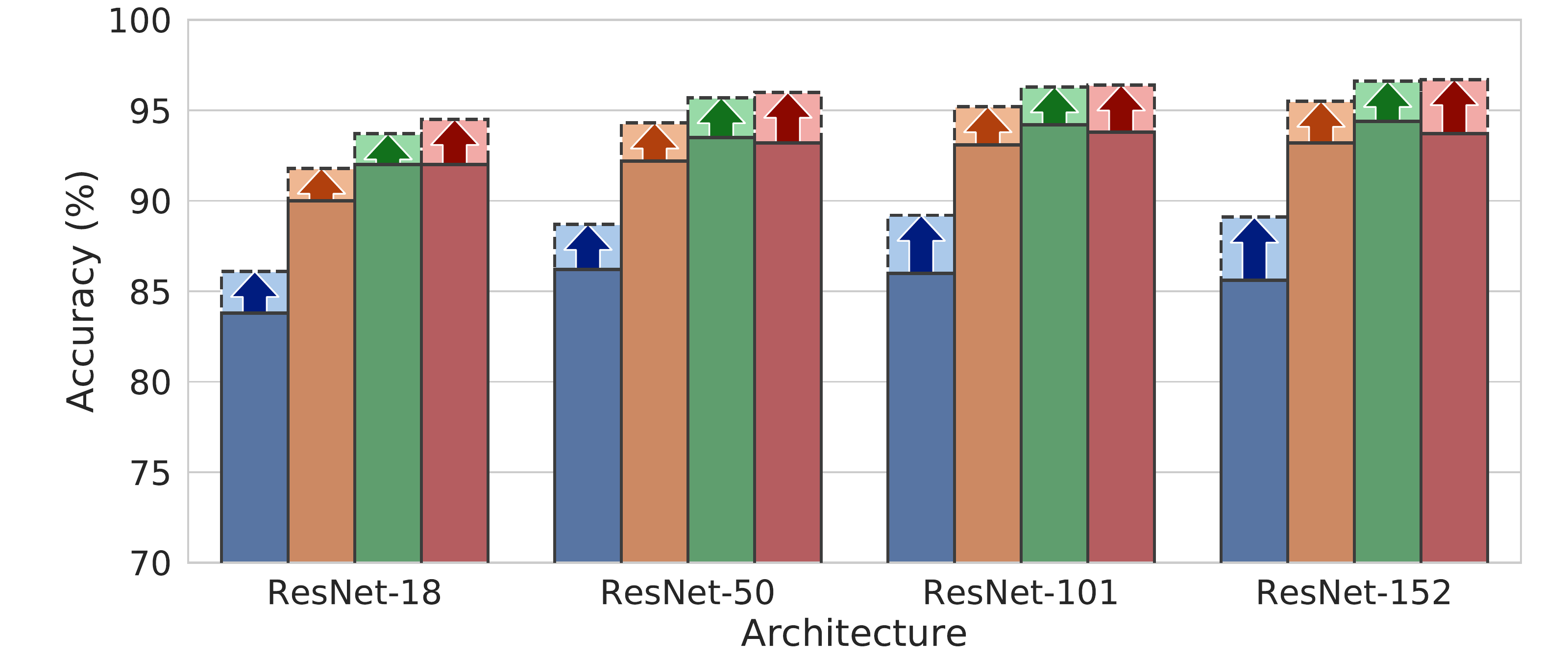}
\vspace{-12pt}
\cutfigurecaptionup
\caption{CIFAR-10}
\cutfigurecaptiondown
\label{fig:arch_epoch_cifar10}
\end{subfigure}
\hfill
\begin{subfigure}[h]{\barrightwidth}
\includegraphics[width=\linewidth]{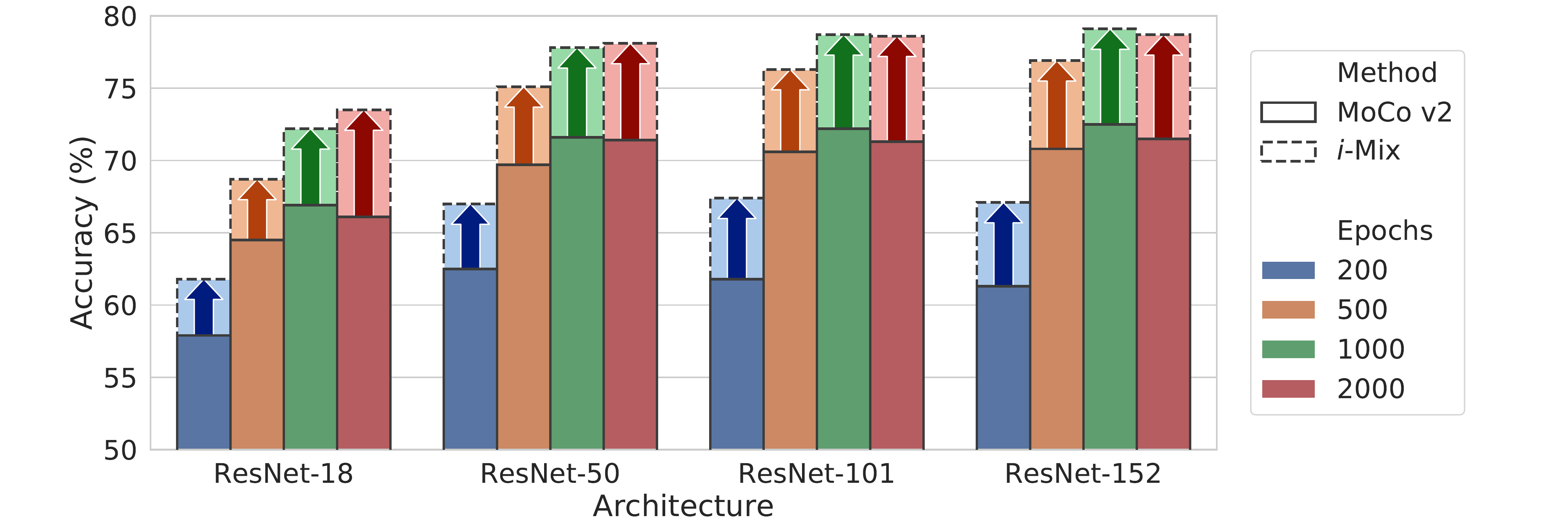}
\vspace{-12pt}
\cutfigurecaptionup
\caption{CIFAR-100}
\cutfigurecaptiondown
\label{fig:arch_epoch_cifar100}
\end{subfigure}
\caption{
Comparison of performance gains by applying {\imix} to MoCo v2 with different model sizes and number of epochs on CIFAR-10 and 100.
%
%
}
\label{fig:arch_epoch}
\end{figure}
%
\begin{figure}[t]
\centering
\begin{minipage}[h]{\tftablewidth}
\centering
\resizebox{\linewidth}{!}{%
\begin{tabular}{cccc}
\toprule
Domain & Dataset & MoCo v2 & + {\imix} \cr
\cmidrule(rl){1-1}\cmidrule(rl){2-2}\cmidrule(rl){3-4}
\multirow{2}{*}{Image}
& ImageNet-100 &
84.1  & \textbf{87.0} \cr
& ImageNet-1k &
70.9  & \textbf{71.3} \cr
\bottomrule
\cr
\toprule
Domain & Dataset & MoCo v2 & + {\imix} \cr
\cmidrule(rl){1-1}\cmidrule(rl){2-2}\cmidrule(rl){3-4}
\multirow{2}{*}{Tabular}
& Higgs100k &
72.1 & \textbf{72.9} \cr
& Higgs1M &
\textbf{74.9} & 74.5 \cr 
\bottomrule
\end{tabular}
}
\vspace{8pt}
\cuttablecaptionup
\captionof{table}{
Comparison of MoCo v2 and {\imix} on large-scale datasets.
}
\cuttablecaptiondown
\label{tb:large}
\end{minipage}
\hfill
\begin{minipage}[h]{\tffigwidth}
\includegraphics[width=\linewidth]{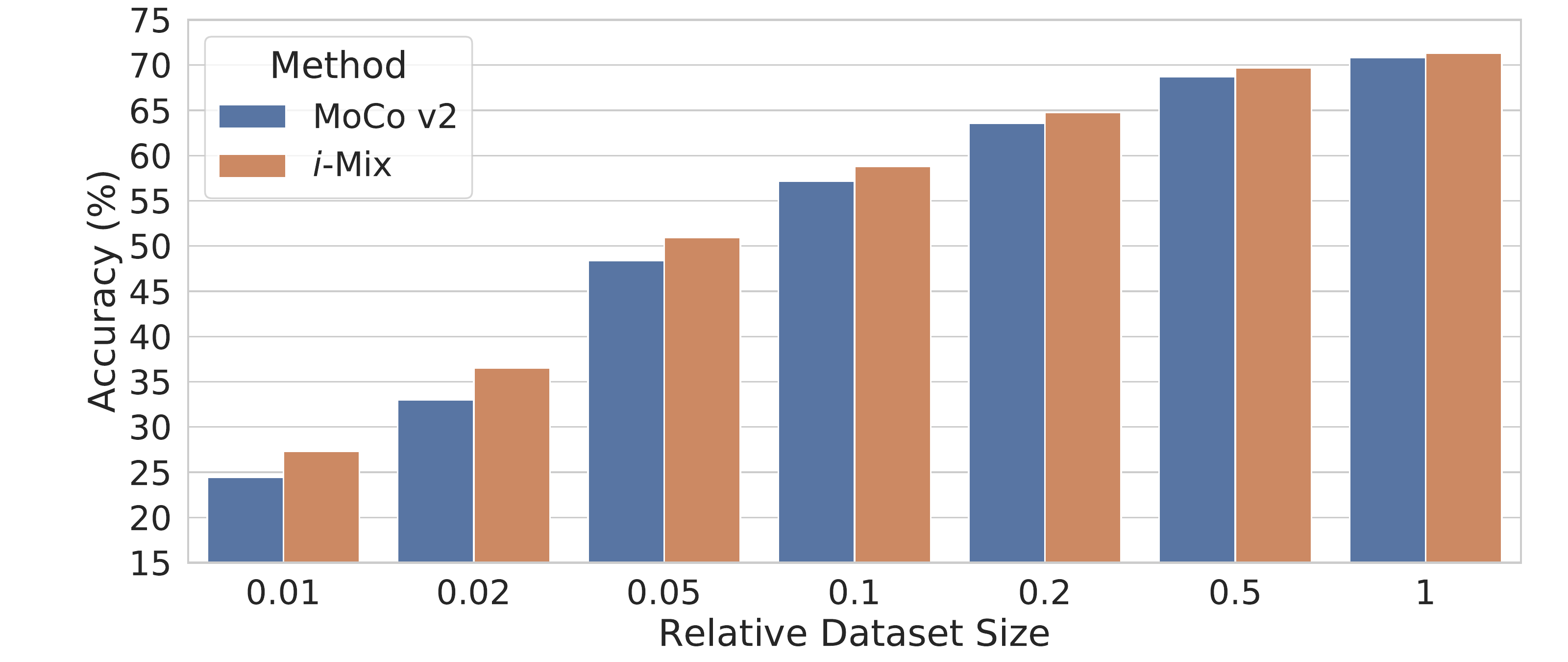}
\vspace{-12pt}
\cutfigurecaptionup
\caption{
Comparison of MoCo v2 and {\imix} trained on the different size of ImageNet.
%
%
}
\cutfigurecaptiondown
\label{fig:dsize}
\end{minipage}
\end{figure}

A better regularization method often benefits from longer training of deeper models, which is more critical when training on a small dataset.
To investigate the regularization effect of {\imix}, we first make a comparison between MoCo v2 and {\imix} by training with different model sizes and number of training epochs on the pretext task.
We train ResNet-18, 50, 101, and 152 models with varying number of training epochs from 200 to 2000.
%
Figure~\ref{fig:arch_epoch} shows the performance of MoCo v2 (solid box) and {\imix} (dashed box).
%
The improvement by applying {\imix} to MoCo v2 is consistent over the different architecture size and the number of training epochs.
%
Deeper models benefit from {\imix}, achieving 96.7\% on CIFAR-10
and 79.1\% on CIFAR-100
when the backbone network is ResNet-152.
%
On the other hand, models trained without {\imix} start to show decrease in performance, possibly due to overfitting to the pretext task when trained longer.
The trend clearly shows that {\imix} results in better representations via improved regularization.

Next, we study the effect of {\imix} with varying dataset sizes for the pretext tasks.
Table~\ref{tb:large} shows the effect of {\imix} on large-scale datasets\footnote{%
Here, ``scale'' corresponds to the amount of data rather than image resolution.
}
from image and tabular domains.
We observe that {\imix} is particularly effective when the amount of training data is reduced,
e.g., ImageNet-100 consists of images from 100 classes, thus has only 10\% of training data compared to ImageNet-1k.
However, the performance gain is reduced when the amount of training data is large.
%
%
%
we further study representations learned with different pretext dataset sizes from 1\% to 100\% of the ImageNet training data in Figure~\ref{fig:dsize}.
Here, different from ImageNet-100, we reduce the amount of data for each class, but maintain the number of classes the same.
%
%
We observe that the performance gain by {\imix} is more significant when the size of the pretext dataset is small.
Our study suggests that {\imix} is effective for regularizing self-supervised representation learning when training from a limited amount of data.
We believe that this is aligned with findings in \citet{zhang2018mixup} for MixUp in supervised learning.
Finally, when a large-scale unlabeled dataset is available, we expect {\imix} would still be useful in obtaining better representations when trained longer with deeper and larger models.

\subsection{Contrastive Learning without Domain-Specific Data Augmentation}
\label{sec:aug}

\begin{table}[t]
\centering
\begingroup
\renewcommand{\thempfootnote}{\fnsymbol{mpfootnote}}
\begin{minipage}[h]{\linewidth}
\centering
\resizebox{\linewidth}{!}{%
\begin{tabular}{ccccccccccccc}
\toprule
\multirow{2}{*}{Aug} & \multicolumn{2}{c}{CIFAR-10} & \multicolumn{2}{c}{CIFAR-100} & \multicolumn{2}{c}{Speech Commands} & \multicolumn{2}{c}{CovType} & \multicolumn{2}{c}{Higgs100k} & \multicolumn{2}{c}{Higgs1M} \cr
\cmidrule(rl){2-3}\cmidrule(rl){4-5}\cmidrule(rl){6-7}\cmidrule(rl){8-9}\cmidrule(rl){10-11}\cmidrule(rl){12-13}
 & MoCo v2 & + {\imix}\footnote{{\inputmix} is applied when no other data augmentations are used.\label{fn:inputmix}} & MoCo v2 & + {\imix}\textsuperscript{\ref{fn:inputmix}} & MoCo v2 & + {\imix} & MoCo v2 & + {\imix} & MoCo v2 & + {\imix} & MoCo v2 & + {\imix} \cr
\cmidrule(rl){1-1}\cmidrule(rl){2-3}\cmidrule(rl){4-5}\cmidrule(rl){6-7}\cmidrule(rl){8-9}\cmidrule(rl){10-11}\cmidrule(rl){12-13}
-      &
{47.7} \scriptsize{$\pm$ 1.3} & \textbf{83.4} \scriptsize{$\pm$ 0.4} &
{24.7} \scriptsize{$\pm$ 0.7} & \textbf{54.0} \scriptsize{$\pm$ 0.5} &
{76.9} \scriptsize{$\pm$ 1.7} & \textbf{92.8} \scriptsize{$\pm$ 0.5} &
{69.6} \scriptsize{$\pm$ 0.3} & \textbf{73.1} \scriptsize{$\pm$ 0.1} &
64.2 & \textbf{71.8} & 65.5 & \textbf{72.9} \cr
\cmark &
{93.5} \scriptsize{$\pm$ 0.2} & \textbf{96.1} \scriptsize{$\pm$ 0.1} &
{71.6} \scriptsize{$\pm$ 0.1} & \textbf{78.1} \scriptsize{$\pm$ 0.3} &
{96.3} \scriptsize{$\pm$ 0.1} & \textbf{98.4} \scriptsize{$\pm$ 0.0} &
{70.5} \scriptsize{$\pm$ 0.2} & \textbf{73.1} \scriptsize{$\pm$ 0.1} &
72.1 & \textbf{72.9} & \textbf{74.9} & 74.5 \cr
\bottomrule
\end{tabular}
}
\cuttablecaptionup
\caption{
Comparison of MoCo v2 and {\imix} with and without data augmentations.
}
\cuttablecaptiondown
\label{tb:aug}
\end{minipage}
\endgroup
\end{table}

\begin{table}[t]
\centering
\begin{subtable}[h]{0.6\linewidth}
\centering
\scalebox{0.85}{%
\begin{tabular}{ccccc}
\toprule
Pretext & \multicolumn{2}{c}{CIFAR-10} & \multicolumn{2}{c}{CIFAR-100} \cr
\cmidrule(rl){1-1}\cmidrule(rl){2-3}\cmidrule(rl){4-5}
Downstream & MoCo v2 & + {\imix} & MoCo v2 & + {\imix} \cr
\cmidrule(rl){1-1}\cmidrule(rl){2-3}\cmidrule(rl){4-5}
CIFAR-10 & {93.5} \scriptsize{$\pm$ 0.2} & \textbf{96.1} \scriptsize{$\pm$ 0.1} & {85.9} \scriptsize{$\pm$ 0.3} & \textbf{90.0} \scriptsize{$\pm$ 0.4} \cr
CIFAR-100 & {64.1} \scriptsize{$\pm$ 0.4} & \textbf{70.8} \scriptsize{$\pm$ 0.4} & {71.6} \scriptsize{$\pm$ 0.1} & \textbf{78.1} \scriptsize{$\pm$ 0.3} \cr
\bottomrule
\end{tabular}
}
\caption{CIFAR-10 and 100 as the pretext dataset}
\end{subtable}
%
%
\begin{subtable}[h]{0.36\linewidth}
\centering
\scalebox{0.85}{%
\begin{tabular}{ccc}
\toprule
\multirow{2}{*}{\parbox{0.35\linewidth}{\centering VOC Object\\Detection}} & \multicolumn{2}{c}{ImageNet} \cr
\cmidrule(rl){2-3}
 & MoCo v2 & + {\imix} \cr
\cmidrule(rl){1-1}\cmidrule(rl){2-3}
AP        & 57.3 \scriptsize{$\pm$ 0.1} & \textbf{57.5} \scriptsize{$\pm$ 0.4} \cr
AP$_{50}$ & 82.5 \scriptsize{$\pm$ 0.2} & \textbf{82.7} \scriptsize{$\pm$ 0.2} \cr
AP$_{75}$ & 63.8 \scriptsize{$\pm$ 0.3} & \textbf{64.2} \scriptsize{$\pm$ 0.7} \cr
\bottomrule
\end{tabular}
}
\caption{ImageNet as the pretext dataset}
\end{subtable}
\cuttablecaptionup
\caption{
Comparison of MoCo v2 and {\imix} in transfer learning.
%
}
\cuttablecaptiondown
\label{tb:transfer}
\end{table}

Data augmentations play a key role in contrastive representation learning,
and therefore it raises a question when applying them to domains with a limited or no knowledge of such augmentations.
In this section, we study the effectiveness of {\imix} as
a domain-agnostic strategy for contrastive representation learning,
which can be adapted to different domains.
%
%
%
Table~\ref{tb:aug} shows the performance of MoCo v2 and {\imix} with and without data augmentations.
We observe significant performance gains with {\imix} when other data augmentations are not applied.
For example,
%
compared to the accuracy of 93.5\% on CIFAR-10 when other data augmentations are applied,
%
contrastive learning achieves 47.7\%
when trained without any data augmentations.
This suggests that data augmentation is an essential part for the success of
contrastive representation learning~\citep{chen2020simple}.
However, {\imix} is able to learn meaningful representations without other data augmentations and achieves
the accuracy of 83.4\% on CIFAR-10.

In Table~\ref{tb:aug}, {\inputmix} is applied together with {\imix} to further improve the performance on image datasets.
For each principal data, we mix two auxiliary data, with mixing coefficients ($0.5 \lambda_{1} \,{+}\, 0.5$, $0.5 \lambda_{2}$, $0.5 \lambda_{3}$), where $\lambda_{1}, \lambda_{2}, \lambda_{3} \,{\sim}\, \text{Dirichlet}(1,1,1)$.\footnote{%
This guarantees that the mixing coefficient for the principal data is larger than 0.5 to prevent from training with noisy labels.
Note that \citet{beckham2019adversarial} also sampled mixing coefficients from the Dirichlet distribution for mixing more than two data.
}
In the above example, while {\imix} is better than baselines, adding {\inputmix} further improves the performance of {\imix}, i.e., from 75.1\% to 83.4\% on CIFAR-10, and from 50.7\% to 54.0\% on CIFAR-100.
This confirms that {\inputmix} can further improve the performance when domain-specific data augmentations are not available, as discussed in Section~\ref{sec:inputmix}.

Moreover, we verify its effectiveness on other domains beyond the image domain.
%
For example, the performance improves from 76.9\% to 92.8\% on the Speech Commands dataset when we assume no other data augmentations are available.
We also observe consistent improvements in accuracy for tabular datasets, even when the training dataset size is large. 
Although the domain knowledge for data augmentations is important to achieve state-of-the-art results,
our demonstration shows the potential of {\imix} to be used for a wide range of application domains where domain knowledge is particularly limited.
\subsection{Transferability of \texorpdfstring{\imix}{\imixtext}}
\label{sec:transfer}

In this section, we show the improved transferability of the representations learned with {\imix}.
The results are provided in Table~\ref{tb:transfer}.
First,
we train linear classifiers with downstream datasets different from the pretext dataset used to train backbone networks and evaluate their performance,
e.g., CIFAR-10 as pretext and CIFAR-100 as downstream datasets or vice versa.
We observe consistent performance gains when learned representations from one dataset are evaluated on classification tasks of another dataset.
Next,
we transfer representations trained on ImageNet to the PASCAL VOC object detection task~\citep{everingham2010pascal}.
We follow the settings in prior works~\citep{he2020momentum,chen2020improved}:
the parameters of the pre-trained ResNet-50 are transferred to a Faster R-CNN detector with the ResNet50-C4 backbone~\citep{ren2015faster}, and fine-tuned end-to-end on the VOC 07+12 trainval dataset and evaluated on the VOC 07 test dataset.
We report the average precision (AP) averaged over IoU thresholds between 50\% to 95\% at a step of 5\%, and
AP$_{50}$ and AP$_{75}$, which are AP values when IoU threshold is 50\% and 75\%, respectively.
Similar to Table~\ref{tb:large}, we observe small but consistent performance gains in all metrics.
Those results confirm that {\imix} improves the quality of learned representations, such that performances on downstream tasks are improved.


\section{Conclusion} \label{sec:conclusion}

We propose {\imix}, a domain-agnostic regularization strategy applicable to a class of self-supervised learning.
The key idea of {\imix} is to introduce a virtual label to each data instance, and mix both inputs and the corresponding virtual labels.
We show that {\imix} is applicable to state-of-the-art self-supervised representation learning methods
including SimCLR, MoCo, and BYOL,
which consistently improves the performance in a variety of settings and domains.
Our experimental results indicate that {\imix} is particularly effective when the training dataset size is small or data augmentation is not available, each of which are prevalent in practice.

\bibliography{ref}
\bibliographystyle{iclr2021_conference}

\newpage

\appendix
\numberwithin{table}{section}
\numberwithin{figure}{section}
\numberwithin{equation}{section}

\ifdefined\supp 
\newcounter{equationmain}
\newenvironment{alignmain}[1][]
{\align\setcounter{equationmain}{#1}}
{\tag{\theequationmain}\endalign}
\fi 

\section{More Applications of \texorpdfstring{\imix}{\imixtext}}
\label{sec:imix_more}

In this section, we introduce more variations of {\imix}.
For conciseness, we use $v_{i}$ to denote virtual labels for different methods.
We make the definition of $v_{i}$ for each application clear.
\ifdefined\supp 
Here we restate the formulation of SimCLR~\citep{chen2020simple}:
\begin{alignmain}[2]
\ell_{\text{SimCLR}} ( x_{i} ; \mathcal{B} ) =
-\log \frac{ \exp \big( s(f_{i}, f_{(N+i) \bmod 2N}) / \tau \big) }
{ \sum_{k=1, k \neq i}^{2N} \exp \big( s(f_{i}, f_{k}) / \tau \big) }.
\label{eq:simclr}
\end{alignmain}
\fi 

\subsection{\texorpdfstring{\imix}{\imixtext} for SimCLR}
\label{sec:simclr_imix}

For each anchor, SimCLR takes other anchors as negative samples such that the virtual labels must be extended.
Let $x_{N+i} \,{=}\, \tilde{x}_{i}$ for conciseness, and $v_{i} \,{\in}\, \{0,1\}^{2N}$ be the virtual label indicating the positive sample of each anchor, where $v_{i,N+i} = 1$ and $v_{i,j \neq N+i} \,{=}\, 0$.
%
%
Note that $v_{i,i} \,{=}\, 0$ because the anchor itself is not counted as a positive sample.
Then, Eq.~\eqref{eq:simclr} can be represented in the form of the cross-entropy loss:
\begin{align}
\ell_{\text{SimCLR}} ( x_{i}, v_{i} ; \mathcal{B} ) =
-\sum_{n=1}^{2N} v_{i,n}
\log \frac{ \exp \big( s(f_{i}, f_{n}) / \tau \big) }
{ \sum_{k=1, k \neq i}^{2N} \exp \big( s(f_{i}, f_{k}) / \tau \big) }.
\label{eq:simclr_v}
\end{align}
The application of {\imix} to SimCLR is straightforward:
for two data instances $(x_{i}, v_{i})$, $(x_{j}, v_{j})$ and a batch of data $\mathcal{B} = \{x_{i}\}_{i=1}^{2N}$, the {\imix} loss is defined as follows:\footnote{The $j$-th data can be excluded from the negative samples, but it does not result in a significant difference.}
\begin{align}
\ell_{\text{SimCLR}}^{\text{\imix}} \big( (x_{i}, v_{i}), (x_{j}, v_{j}) ; \mathcal{B}, \lambda \big) =
\ell_{\text{SimCLR}}( \lambda x_{i} + (1 - \lambda) x_{j}, \lambda v_{i} + (1 - \lambda) v_{j} ; \mathcal{B} ).
\label{eq:simclr_imix}
\end{align}

Note that only the input data of Eq.~\eqref{eq:simclr_imix} is mixed, such that $f_{i}$ in Eq.~\eqref{eq:simclr_v} is an embedding vector of the mixed data while the other $f_{n}$'s are the ones of clean data.
Because both clean and mixed data
need to be fed to the network $f$, {\imix} for SimCLR
requires twice more memory and training time compared to
%
SimCLR when the same batch size is used.

\subsection{\texorpdfstring{\imix}{\imixtext} for Supervised Contrastive Learning}
\label{sec:supcon_imix}

Supervised contrastive learning has recently shown to be effective for supervised representation learning and it often outperforms the standard end-to-end supervised classifier learning~\citep{khosla2020supervised}.
Suppose an one-hot label $y_{i} \in \{0,1\}^{C}$ is assigned to a data $x_{i}$, where $C$ is the number of classes.
Let $x_{N+i} \,{=}\, \tilde{x}_{i}$ and $y_{N+i} \,{=}\, y_{i}$ for conciseness.
For a batch of data pairs and their labels $\mathcal{B} \,{=}\, \{(x_{i}, y_{i})\}_{i=1}^{2N}$,
let $v_{i} \,{\in}\, \{0,1\}^{2N}$ be the virtual label indicating the positive samples of each anchor,
where $v_{i,j} = 1$ if $y_{i} \,{=}\, y_{j \neq i}$,
and otherwise $v_{i,j} \,{=}\, 0$.
Intuitively, $\sum_{j=1}^{2N} v_{i,j} = 2N_{y_{i}} - 1$ where $N_{y_{i}}$ is the number of data with the label $y_{i}$.
Then, the supervised learning version of the SimCLR (SupCLR) loss function is written as follows:
\begin{align}
\ell_{\text{SupCLR}} ( x_{i}, v_{i} ; \mathcal{B} ) =
-\frac{1}{2N_{y_{i}} - 1}
\sum_{n=1}^{2N} v_{i,n}
\log \frac{ \exp \big( s(f_{i}, f_{n}) / \tau \big) }
{ \sum_{k=1, k \neq i}^{2N} \exp \big( s(f_{i}, f_{k}) / \tau \big) }.
\label{eq:supcon}
\end{align}
The application of {\imix} to SupCLR is straightforward:
for two data instances $(x_{i}, v_{i})$, $(x_{j}, v_{j})$ and a batch of data $\mathcal{B} = \{x_{i}\}_{i=1}^{2N}$, the {\imix} loss is defined as follows:
\begin{align}
\ell_{\text{SupCLR}}^{\text{{\imix}}} \big( (x_{i}, v_{i}), (x_{j}, v_{j}) ; \mathcal{B}, \lambda \big) =
\ell_{\text{SupCLR}}( \lambda x_{i} + (1 - \lambda) x_{j}, \lambda v_{i} + (1 - \lambda) v_{j} ; \mathcal{B} ).
\label{eq:supcon_mix}
\end{align}

\subsection{\texorpdfstring{\imix}{\imixtext} for N-Pair Supervised Contrastive Learning}
\label{sec:npair_supcon_mix}

Note that {\imix} in Eq.~\eqref{eq:supcon_mix} is not as efficient as SupCLR in Eq.~\eqref{eq:supcon} due to the same reason in the case of SimCLR.
To overcome this, we reformulate SupCLR in the form of the N-pair loss~\citep{sohn2016improved}.
Suppose an one-hot label $y_{i} \in \{0,1\}^{C}$ is assigned to a data $x_{i}$, where $C$ is the number of classes.
For a batch of data pairs and their labels $\mathcal{B} \,{=}\, \{(x_{i}, \tilde{x}_{i}, y_{i})\}_{i=1}^{N}$,
let $v_{i} \,{\in}\, \{0,1\}^{N}$ be the virtual label indicating the positive samples of each anchor,
where $v_{i,j} = 1$ if $y_{i} \,{=}\, y_{j \neq i}$,
and otherwise $v_{i,j} \,{=}\, 0$.
Then, the supervised version of the N-pair (Sup-N-pair) contrastive loss function is written as follows:
\begin{align}
\ell_{\text{Sup-N-pair}} ( x_{i}, v_{i} ; \mathcal{B} ) =
-\frac{1}{N_{y_{i}}}
\sum_{n=1}^{N} v_{i,n}
\log \frac{ \exp \big( s(f_{i}, \tilde{f}_{n}) / \tau \big) }
{ \sum_{k=1}^{N} \exp \big( s(f_{i}, \tilde{f}_{k}) / \tau \big) }.
\label{eq:supcon_npair}
\end{align}
Then, the {\imix} loss for Sup-N-pair is defined as follows:
\begin{align}
\ell_{\text{Sup-N-pair}}^{\text{{\imix}}} \big( (x_{i}, v_{i}), (x_{j}, v_{j}) ; \mathcal{B}, \lambda \big) =
\ell_{\text{Sup-N-pair}}( \lambda x_{i} + (1 - \lambda) x_{j}, \lambda v_{i} + (1 - \lambda) v_{j} ; \mathcal{B} ).
\label{eq:supcon_npair_mix}
\end{align}





\section{Proof of the linearity of losses with respect to virtual labels}
\label{sec:linear}

\paragraph{Cross-entropy loss.}
The loss used in contrastive representation learning works, which is often referred to as InfoNCE~\citep{oord2018representation}, can be represented in the form of the cross-entropy loss as we showed for N-pair contrastive learning, SimCLR~\citep{chen2020simple}, and MoCo~\citep{he2020momentum}.
Here we provide an example in the case of N-pair contrastive learning.
Let $f_{ij}^{\lambda} \,{=}\, f(\lambda x_{i} \,{+}\, (1 \,{-}\, \lambda) x_{j})$ for conciseness.
\begin{align}
&\ell_{\text{N-pair}}^{\text{\imix}} \big( (x_{i}, v_{i}), (x_{j}, v_{j}) ; \mathcal{B}, \lambda \big) =
\ell_{\text{N-pair}}( \lambda x_{i} + (1 - \lambda) x_{j}, \lambda v_{i} + (1 - \lambda) v_{j} ; \mathcal{B}) \nonumber \\
&= -\sum_{n=1}^{N} (\lambda v_{i,n} + (1 - \lambda) v_{j,n})
\log \frac{ \exp \big( s(f_{ij}^{\lambda}, \tilde{f}_{n}) / \tau \big) }
{ \sum_{k=1}^{N} \exp \big( s(f_{ij}^{\lambda}, \tilde{f}_{k}) / \tau \big) } \nonumber \\
&= - \lambda \sum_{n=1}^{N} v_{i,n}
\log \frac{ \exp \big( s(f_{ij}^{\lambda}, \tilde{f}_{n}) / \tau \big) }
{ \sum_{k=1}^{N} \exp \big( s(f_{ij}^{\lambda}, \tilde{f}_{k}) / \tau \big) }
- (1 - \lambda) \sum_{n=1}^{N} v_{j,n}
\log \frac{ \exp \big( s(f_{ij}^{\lambda}, \tilde{f}_{n}) / \tau \big) }
{ \sum_{k=1}^{N} \exp \big( s(f_{ij}^{\lambda}, \tilde{f}_{k}) / \tau \big) } \nonumber \\
&= \lambda \ell_{\text{N-pair}}( \lambda x_{i} + (1 - \lambda) x_{j}, v_{i} ; \mathcal{B}) +
(1 - \lambda) \ell_{\text{N-pair}}( \lambda x_{i} + (1 - \lambda) x_{j}, v_{j} ; \mathcal{B}).
\label{eq:npair_imix_long}
\end{align}


\paragraph{L2 loss between L2-normalized feature vectors.} \hspace{0pt}
The BYOL~\citep{grill2020bootstrap} loss is in this type.
Let $\tilde{F} \,{=}\, [\tilde{f}_{1} / \lVert \tilde{f}_{1} \rVert, {\dots}, \tilde{f}_{N} / \lVert \tilde{f}_{N} \rVert] \,{\in}\, \mathbb{R}^{D {\times} N}$ such that $\tilde{f}_{i} / \lVert \tilde{f}_{i} \rVert \,{=}\, \tilde{F} v_{i}$, and
$\bar{g} \,{=}\, g(f( \lambda x_{i} \,{+}\, (1 \,{-}\, \lambda) x_{j} )) / \lVert g(f( \lambda x_{i} \,{+}\, (1 \,{-}\, \lambda) x_{j} )) \rVert$ for conciseness.
\begin{align}
&\ell_{\text{BYOL}}^{\text{\imix}} \big( (x_{i}, v_{i}), (x_{j}, v_{j}) ; \mathcal{B}, \lambda \big)
= \ell_{\text{BYOL}}( \lambda x_{i} + (1 - \lambda) x_{j}, \lambda v_{i} + (1 - \lambda) v_{j} ) \nonumber \\
&= \left\lVert \bar{g} - \tilde{F} ( \lambda v_{i} + (1 - \lambda) v_{j} ) \right\rVert^{2}
= \left\lVert \bar{g} - \left( \lambda \tilde{F} v_{i} + (1 - \lambda) \tilde{F} v_{j} \right) \right\rVert^{2} \nonumber \\
&= 1
- 2 \cdot \bar{g}^{\top} \left( \lambda \tilde{F} v_{i} + (1 - \lambda) \tilde{F} v_{j} \right)
+ \left\lVert \lambda \tilde{F} v_{i} + (1 - \lambda) \tilde{F} v_{j} \right\rVert^{2}
\nonumber \\
&= 2
- 2 \cdot \bar{g}^{\top} \left( \lambda \tilde{F} v_{i} + (1 - \lambda) \tilde{F} v_{j} \right)
+ \text{const} \nonumber \\
&= \lambda \lVert \bar{g} - \tilde{F} v_{i} \rVert^{2}
+ (1 - \lambda) \lVert \bar{g} - \tilde{F} v_{j} \rVert^{2}
+ \text{const} \nonumber \\
&= \lambda \ell_{\text{BYOL}}( \lambda x_{i} + (1 - \lambda) x_{j}, v_{i} ; \mathcal{B})
+ (1 - \lambda) \ell_{\text{BYOL}}( \lambda x_{i} + (1 - \lambda) x_{j}, v_{j} ; \mathcal{B})
+ \text{const}.
\label{eq:byol_imix_long}
\end{align}
Because $\tilde{F}$ is not backpropagated, it can be considered as a constant.

\section{More on Experiments}
\label{sec:exp_more}

We describe details of the experimental settings and more experimental results.
For additional experiments below, we adapted the code for supervised contrastive learning~\citep{khosla2020supervised}.\footnote{\url{https://github.com/HobbitLong/SupContrast}}

\subsection{Setup}
\label{sec:setup_more}

In this section, we describe details of the experimental settings.
Note that the learning rate is scaled by the batch size~\citep{goyal2017accurate}:
$\text{ScaledLearningRate} \,{=}\, \text{LearningRate} \times \text{BatchSize} / 256$.


\paragraph{Image.}
The experiments on CIFAR-10 and 100~\citep{krizhevsky2009learning} and ImageNet~\citep{deng2009imagenet} are conducted in two stages:
following \citet{chen2020simple},
%
the convolutional neural network (CNN) part of ResNet-50~\citep{he2016deep}\footnote{%
For small resolution data from CIFAR and Speech Commands, we replaced the kernal, stride, and padding size from (7,2,3) to (3,1,1) in the first convolutional layer, and removed the first max pooling layer, following \citet{chen2020simple}.
}
followed by the two-layer multilayer perceptron (MLP) projection head (output dimensions are 2048 and 128, respectively) is trained on the unlabeled pretext dataset
with a batch size of 256 (i.e., 512 augmented data) with the stochastic gradient descent (SGD) optimizer with a momentum of 0.9 over up to 4000 epochs.
BYOL has an additional prediction head (output dimensions are the same with the projection head), which follows the projection head, only for the model updated by gradient.
10 epochs of warmup with a linear schedule to an initial learning rate of 0.125, followed by the cosine learning rate schedule~\citep{loshchilov2017sgdr} is used.
We use the weight decay of 0.0001 for the first stage.
For ImageNet, we use the same hyperparameters except that the batch size is 512 and the initial learning rate is 0.03.

Then, the head of the CNN is replaced with a linear classifier, and only the linear classifier is trained with the labeled downstream dataset.
%
For the second stage, we use a batch size of 256 with the SGD optimizer with a momentum of 0.9 and an initial learning rate chosen among \{1, 3, 5, 10, 30, 50, 70\} over 100 epochs, where the learning rate is decayed by 0.2 after 80, 90, 95 epochs.
No weight decay is used at the second stage.
The quality of representation is evaluated by the top-1 accuracy on the downstream task.
We sample a single mixing coefficient $\lambda \,{\sim}\, \text{Beta}(1,1)$ for each training batch.
%
The temperature is set to $\tau \,{=}\, 0.2$.
Note that the optimal distribution of $\lambda$ and the optimal value of $\tau$ varies over different architectures, methods, and datasets, but the choices above result in a reasonably good performance.
The memory bank size of MoCo is 65536 for ImageNet and 4096 for other datasets, and the momentum for the exponential moving average (EMA) update is 0.999 for MoCo and BYOL.
We do not symmetrize the BYOL loss, as it does not significantly improve the performance while increasing computational complexity.
%

For data augmentation, we follow \citet{chen2020simple}:
We apply a set of data augmentations randomly in sequence including
resized cropping~\citep{szegedy2015going},
horizontal flipping with a probability of 0.5,
color jittering,\footnote{Specifically, brightness, contrast, and saturation are scaled by a factor uniformly sampled from $[0.6,1.4]$ at random, and hue is rotated in the HSV space by a factor uniformly sampled from $[-0.1,0.1]$ at random.} and
gray scaling with a probability of 0.2.
A Gaussian blurring with $\sigma \in [0.1, 2]$ and kernel size of 10\% of the image height/width is applied for ImageNet.
For evaluation on downstream tasks, we apply padded cropping with the pad size of 4 and horizontal flipping for CIFAR-10 and 100, and resized cropping and horizontal flipping for ImageNet.


\paragraph{Speech.}
In the experiments on Speech Commands~\citep{warden2018speech},
%
the network is the same with the image domain experiments, except that the number of input channels is one instead of three.
The temperature is set to $\tau \,{=}\, 0.5$ for the standard setting and $\tau \,{=}\, 0.2$ for the no augmentation setting.
10\% of silence data (all zero) are added when training.
At the first stage, the model is trained with the SGD optimizer with a momentum of 0.9 and an initial learning rate of 0.125 over 500 epochs, where the learning rate decays by 0.1 after 300 and 400 epochs and the weight decay is 0.0001.
The other settings are the same with the experiments on CIFAR.
%
%
%
%

For data augmentation,\footnote{\url{https://github.com/tugstugi/pytorch-speech-commands}} we apply a set of data augmentations randomly in sequence including
changing amplitude, speed, and pitch in time domain,
stretching, time shifting, and adding background noise in frequency domain.
Each data augmentation is applied with a probability of 0.5.
Augmented data are then transformed to the mel spectogram in the size of 32\,$\times$\,32.

\paragraph{Tabular.}
In the experiments on CovType and Higgs~\citep{asuncion2007uci},
we take a five-layer MLP with batch normalization as a backbone network.
The output dimensions of layers are (2048-2048-4096-4096-8192), where all layers have batch normalization followed by ReLU except for the last layer.
The last layer activation is maxout~\citep{goodfellow2013maxout} with 4 sets, such that the output dimension is 2048.
On top of this five-layer MLP, we attach two-layer MLP (2048-128) as a projection head.
We sample a single mixing coefficient $\lambda \,{\sim}\, \text{Beta}(\alpha, \alpha)$ for each training batch, where $\alpha \,{=}\, 2$ for CovType and Higgs100k, and $\alpha \,{=}\, 1$ for Higgs1M.
The temperature is set to $\tau \,{=}\, 0.1$.
%
%
The other settings are the same with the experiments on CIFAR, except that the batch size is 512 and the number of training epochs is 500.
At the second stage, the MLP head is replaced with a linear classifier.
For Higgs, the classifier is computed by linear regression from the feature matrix obtained without data augmentation to the label matrix using the pseudoinverse.
Since the prior knowledge on tabular data is very limited,
%
only the masking noise with a probability of 0.2 is considered as a data augmentation.


\subsection{Variations of \texorpdfstring{\imix}{\imixtext}}
\label{sec:var}

\begin{table}[t]
\centering
\resizebox{\linewidth}{!}{%
\begin{tabular}{cccccccc}
\toprule
\multirow{2}{*}{Pretext} & \multirow{2}{*}{Downstream} & \multicolumn{3}{c}{N-pair} & \multicolumn{3}{c}{SimCLR} \cr
\cmidrule(rl){3-5}\cmidrule(rl){6-8}
& & Vanilla & {\imixup} & {\icutmix} & Vanilla & {\imixup} & {\icutmix} \cr
\cmidrule(rl){1-1}\cmidrule(rl){2-2}\cmidrule(rl){3-5}\cmidrule(rl){6-8}
\multirow{2}{*}{CIFAR-10} & CIFAR-10 &
92.4 \scriptsize{$\pm$ 0.1} & \textbf{94.8} \scriptsize{$\pm$ 0.2} & 94.7 \scriptsize{$\pm$ 0.1} &
92.5 \scriptsize{$\pm$ 0.1} & \textbf{94.8} \scriptsize{$\pm$ 0.2} & \textbf{94.8} \scriptsize{$\pm$ 0.2} \cr
& CIFAR-100 &
60.2 \scriptsize{$\pm$ 0.3} & \textbf{63.3} \scriptsize{$\pm$ 0.2} & 61.5 \scriptsize{$\pm$ 0.2} &
60.0 \scriptsize{$\pm$ 0.2} & \textbf{61.4} \scriptsize{$\pm$ 1.0} & 57.1 \scriptsize{$\pm$ 0.4} \cr
\cmidrule(rl){1-1}\cmidrule(rl){2-2}\cmidrule(rl){3-5}\cmidrule(rl){6-8}
\multirow{2}{*}{CIFAR-100} & CIFAR-10 &
84.4 \scriptsize{$\pm$ 0.2} & \textbf{86.2} \scriptsize{$\pm$ 0.2} & 85.1 \scriptsize{$\pm$ 0.2} &
84.4 \scriptsize{$\pm$ 0.2} & \textbf{85.2} \scriptsize{$\pm$ 0.3} & 83.7 \scriptsize{$\pm$ 0.6} \cr
& CIFAR-100 &
68.7 \scriptsize{$\pm$ 0.2} & \textbf{72.3} \scriptsize{$\pm$ 0.2} & \textbf{72.3} \scriptsize{$\pm$ 0.4} &
68.7 \scriptsize{$\pm$ 0.2} & \textbf{72.3} \scriptsize{$\pm$ 0.2} & 71.7 \scriptsize{$\pm$ 0.2} \cr
\bottomrule
\end{tabular}
}
\cuttablecaptionup
\caption{
Comparison of N-pair contrastive learning and SimCLR with {\imixup} and {\icutmix} on them with ResNet-50 on CIFAR-10 and 100.
%
We run all experiments for 1000 epochs.
%
%
{\imixup} improves the accuracy on the downstream task regardless of the data distribution shift between the pretext and downstream tasks.
{\icutmix} shows a comparable performance with {\imixup} when the pretext and downstream datasets are the same, but it does not when the data distribution shift occurs.
}
\cuttablecaptiondown
\label{tb:variation}
\end{table}

\begin{table}[t]
\centering
\resizebox{\linewidth}{!}{%
\begin{tabular}{cccccccc}
\toprule
\multirow{2}{*}{Pretext} & \multirow{2}{*}{Downstream} & \multicolumn{3}{c}{Self-Supervised Pretext} & \multicolumn{3}{c}{Supervised Pretext} \cr
\cmidrule(rl){3-5}\cmidrule(rl){6-8}
& & SimCLR & N-pair & + {\imix} & SimCLR & N-pair & + {\imix} \cr
\cmidrule(rl){1-1}\cmidrule(rl){2-2}\cmidrule(rl){3-5}\cmidrule(rl){6-8}
\multirow{2}{*}{CIFAR-10} & CIFAR-10 &
92.5 \scriptsize{$\pm$ 0.1} & 92.4 \scriptsize{$\pm$ 0.1} & \textbf{94.8} \scriptsize{$\pm$ 0.2} &
95.6 \scriptsize{$\pm$ 0.3} & 95.7 \scriptsize{$\pm$ 0.1} & \textbf{97.0} \scriptsize{$\pm$ 0.1} \cr
& CIFAR-100 &
60.0 \scriptsize{$\pm$ 0.2} & 60.2 \scriptsize{$\pm$ 0.3} & \textbf{63.3} \scriptsize{$\pm$ 0.2} &
58.6 \scriptsize{$\pm$ 0.2} & \textbf{58.9} \scriptsize{$\pm$ 0.5} & 57.8 \scriptsize{$\pm$ 0.6} \cr
\cmidrule(rl){1-1}\cmidrule(rl){2-2}\cmidrule(rl){3-5}\cmidrule(rl){6-8}
\multirow{2}{*}{CIFAR-100} & CIFAR-10 &
84.4 \scriptsize{$\pm$ 0.2} & 84.4 \scriptsize{$\pm$ 0.2} & \textbf{86.2} \scriptsize{$\pm$ 0.2} &
86.5 \scriptsize{$\pm$ 0.4} & 86.7 \scriptsize{$\pm$ 0.2} & \textbf{88.7} \scriptsize{$\pm$ 0.2} \cr
& CIFAR-100 &
68.7 \scriptsize{$\pm$ 0.2} & 68.7 \scriptsize{$\pm$ 0.2} & \textbf{72.3} \scriptsize{$\pm$ 0.2} &
74.3 \scriptsize{$\pm$ 0.2} & 74.6 \scriptsize{$\pm$ 0.3} & \textbf{78.4} \scriptsize{$\pm$ 0.2} \cr
\bottomrule
\end{tabular}
}
\cuttablecaptionup
\caption{
Comparison of the N-pair self-supervised and supervised contrastive learning methods and {\imix} on them with ResNet-50 on CIFAR-10 and 100.
We also provide the performance of formulations proposed in prior works:
SimCLR~\citep{chen2020simple} and its supervised version~\citep{khosla2020supervised}.
%
We run all experiments for 1000 epochs.
%
%
{\imix} improves the accuracy on the downstream task regardless of the data distribution shift between the pretext and downstream tasks, except the case that the pretest task has smaller number of classes than that of the downstream task.
The quality of representation depends on the pretext task in terms of the performance of transfer learning:
self-supervised learning is better on CIFAR-10, while supervised learning is better on CIFAR-100.
}
\cuttablecaptiondown
\label{tb:supcon}
\end{table}

We compare the MixUp~\citep{zhang2018mixup} and CutMix~\citep{yun2019cutmix} variation of {\imix} on N-pair contrastive learning and SimCLR.
To distinguish them, we call them {\imixup} and {\icutmix}, respectively.
To be fair with the memory usage in the pretext task stage, we reduce the batch size of {\imixup} and {\icutmix} by half (256 to 128) for SimCLR.
Following the learning rate adjustment strategy in \citet{goyal2017accurate}, we also decrease the learning rate by half (0.125 to 0.0625) when the batch size is reduced.
We note that {\imixup} and {\icutmix} on SimCLR take approximately 2.5 times more training time to achieve the same number of training epochs.
The results are provided in Table~\ref{tb:variation}.
%
We first verify that the N-pair formulation results in no worse performance than that of SimCLR.
This justifies to conduct experiments using the N-pair formulation instead of that of SimCLR, which is simpler and more efficient, especially when applying {\imix}, while not losing the performance.
When pretext and downstream tasks share the training dataset, {\icutmix} often outperforms {\imixup}, though the margin is small.
However, {\icutmix} shows a worse performance in transfer learning.

Table~\ref{tb:supcon} compares the performance of SimCLR, N-pair contrastive learning, and {\imix} on N-pair contrastive learning when the pretext task is self-supervised and supervised contrastive learning.
We confirm that the N-pair formulation results in no worse performance than that of SimCLR in supervised contrastive learning as well.
{\imix} improves the performance of supervised contrastive learning from 95.7\% to 97.0\% on CIFAR-10, similarly to improvement achieved by MixUp for supervised learning where it improves the performance of supervised classifier learning from 95.5\% to 96.6\%.
On the other hand, when the pretext dataset is CIFAR-100, the performance of supervised contrastive learning is not better than that of supervised learning:
MixUp improves the performance of supervised classifier learning from 78.9\% to 82.2\%, and {\imix} improves the performance of supervised contrastive learning from 74.6\% to 78.4\%.

While supervised {\imix} improves the classification accuracy on CIFAR-10 when trained on CIFAR-10, the representation does not transfer well to CIFAR-100, possibly due to overfitting to 10 class classification. 
%
%
%
When pretext dataset is CIFAR-100, supervised contrastive learning shows a better performance than self-supervised contrastive learning regardless of the distribution shift, as it learns sufficiently general representation for linear classifier to work well on CIFAR-10 as well.
%
%

\subsection{Qualitative Embedding Analysis}
\label{sec:qual_embed}

\begin{figure}[t]
\vspace{-20pt}
\centering
\begin{subfigure}[h]{\scatterleftwidth}
\includegraphics[width=\linewidth]{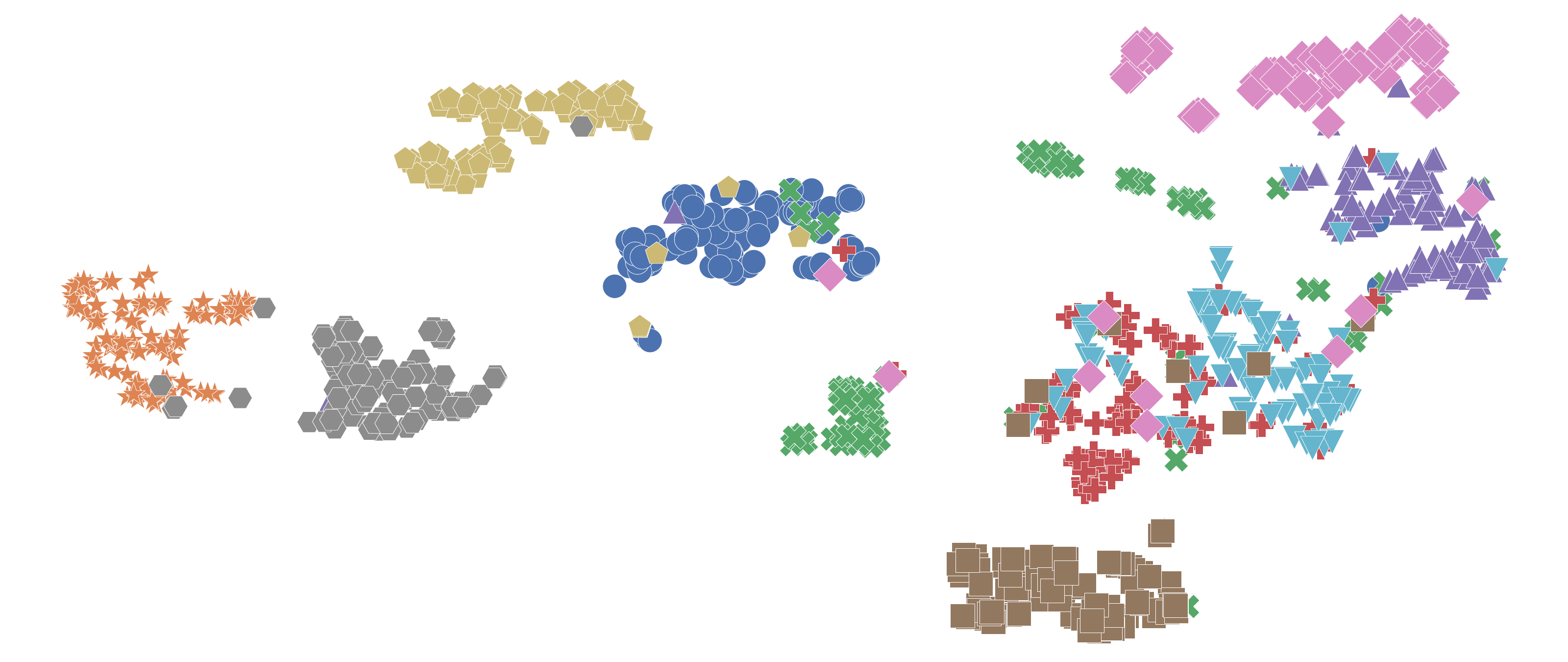}
\caption{Contrastive Learning on CIFAR-10}
\label{fig:tsne_npair_cifar10}
\end{subfigure}
\captionsetup[subfigure]{oneside,margin={0pt,40pt}}
\begin{subfigure}[h]{\scatterrightwidth}
\includegraphics[width=\linewidth]{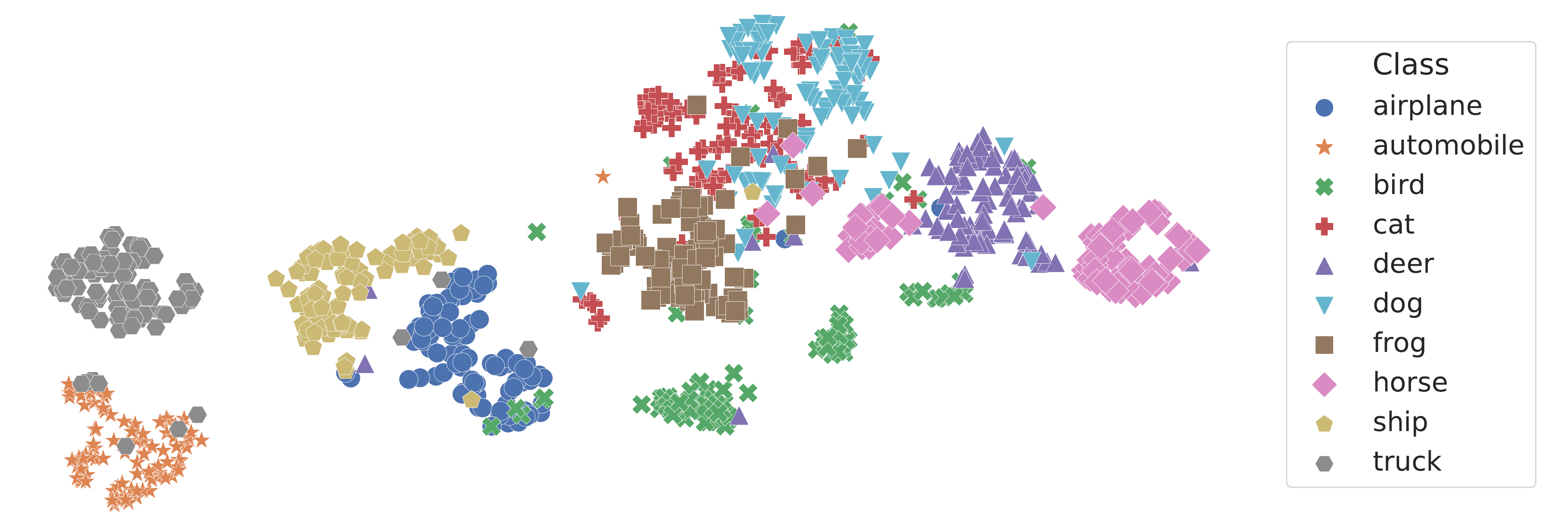}
\caption{{\imix} on CIFAR-10}
\label{fig:tsne_mixup_cifar10}
\end{subfigure} \\
\captionsetup[subfigure]{oneside,margin={0pt,0pt}}
\begin{subfigure}[h]{\scatterleftwidth}
\includegraphics[width=\linewidth]{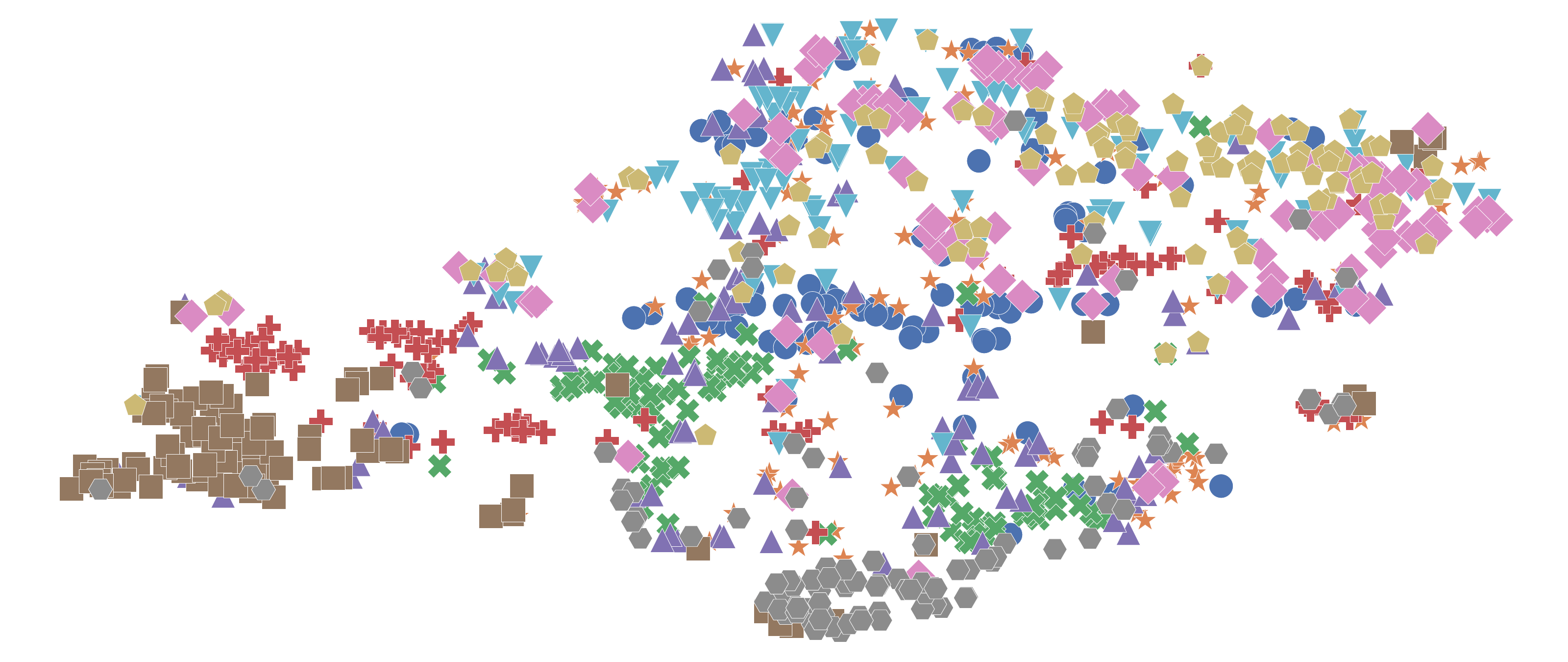}
\caption{Contrastive Learning on CIFAR-100}
\label{fig:tsne_npair_cifar100}
\end{subfigure}
\captionsetup[subfigure]{oneside,margin={0pt,40pt}}
\begin{subfigure}[h]{\scatterrightwidth}
\includegraphics[width=\linewidth]{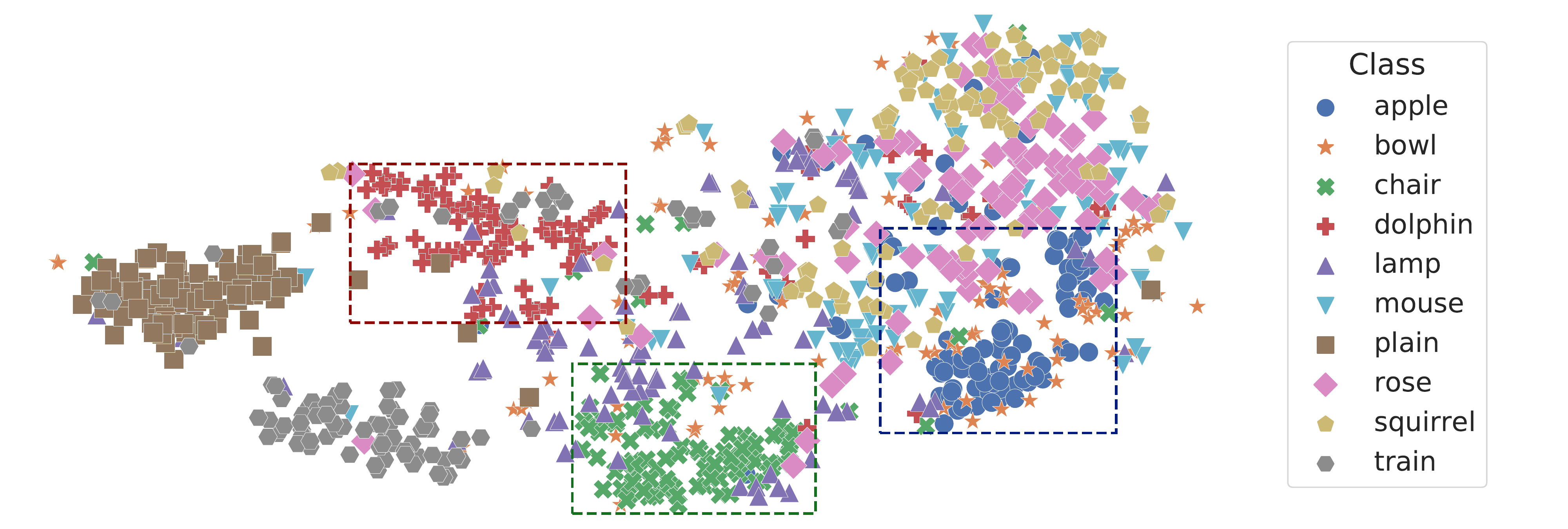}
\caption{{\imix} on CIFAR-100}
\label{fig:tsne_mixup_cifar100}
\end{subfigure} \\
\cutfigurecaptionup
\caption{
t-SNE visualization of embeddings trained by contrastive learning and {\imix} with ResNet-50 on CIFAR-10.
(a,b): Classes are well-clustered in both cases when applied to CIFAR-10.
(c,d): When models are transferred to CIFAR-100, classes are more clustered for {\imix} than contrastive learning, as highlighted in dashed boxes.
We show 10 classes for a better visualization.
}
\cutfigurecaptiondown
\label{fig:tsne}
\end{figure}

Figure~\ref{fig:tsne} visualizes embedding spaces learned by N-pair contrastive learning and {\imix} on CIFAR-10 and 100.
When the downstream dataset is the same with the pretext task, 
both contrastive learning and {\imix} cluster classes well, as shown in  Figure~\ref{fig:tsne_npair_cifar10} and \ref{fig:tsne_mixup_cifar10}.
However, when the downstream task is transferred to CIFAR-100, {\imix} in Figure~\ref{fig:tsne_mixup_cifar100} clusters classes better than contrastive learning in Figure~\ref{fig:tsne_npair_cifar100}.
Specifically, clusters of ``apple,'' ``chair,'' and ``dolphin,'' can be found in Figure~\ref{fig:tsne_mixup_cifar100} while they spread out
in Figure~\ref{fig:tsne_npair_cifar100}.
Also, ``rose'' and ``squirrel'' are more separated in Figure~\ref{fig:tsne_mixup_cifar100} than \ref{fig:tsne_npair_cifar100}.
This shows that the representation learned with {\imix} is more generalizable than vanilla contrastive learning.

\subsection{Quantitative Embedding Analysis}
\label{sec:quan_embed}

\begin{table}[t]
\centering
\begin{tabular}{cccccccc}
\toprule
\multirow{2}{*}{Pretext} & \multirow{2}{*}{Downstream} & \multicolumn{2}{c}{FED ($\times 10^{-4}$) ($\downarrow$)} & \multicolumn{2}{c}{Training Acc (\%) ($\uparrow$)} & \multicolumn{2}{c}{Test Acc (\%) ($\uparrow$)} \cr
\cmidrule(rl){3-4}\cmidrule(rl){5-6}\cmidrule(rl){7-8}
& & N-pair & + {\imix} & N-pair & + {\imix} & N-pair & + {\imix} \cr
\cmidrule(rl){1-1}\cmidrule(rl){2-2}\cmidrule(rl){3-4}\cmidrule(rl){5-6}\cmidrule(rl){7-8}
\multirow{2}{*}{CIFAR-10} & CIFAR-10 &
30.0 & \textbf{16.7} & \textbf{96.1} & \textbf{96.1} & 92.4 & \textbf{94.8} \cr
 & CIFAR-100 &
13.8 & \textbf{7.9} & \textbf{70.7} & 69.5 & 60.2 & \textbf{63.3} \cr
\cmidrule(rl){1-1}\cmidrule(rl){2-2}\cmidrule(rl){3-4}\cmidrule(rl){5-6}\cmidrule(rl){7-8}
\multirow{2}{*}{CIFAR-100} & CIFAR-10 &
15.2 & \textbf{9.7} & 88.1 & \textbf{88.8} & 84.4 & \textbf{86.2} \cr
 & CIFAR-100 &
30.4 & \textbf{13.3} & \textbf{85.6} & 79.0 & 68.7 & \textbf{72.3} \cr
\bottomrule
\end{tabular}
\caption{
Comparison of N-pair contrastive learning and {\imix} with ResNet-50 on CIFAR-10 and 100 in terms of the Fr{\'e}chet embedding distance (FED) between training and test data distribution on the embedding space, and training and test accuracy.
$\uparrow$ ($\downarrow$) indicates that the higher (lower) number is the better.
{\imix} improves contrastive learning in all metrics, which shows that {\imix} is an effective regularization method for the pretext task, such that the learned representation is more generalized.
}
\vspace{-4pt}
\label{tb:fed}
\end{table}

To estimate the quality of representation by the similarity between training and test data distribution,
we measure the Fr{\'e}chet embedding distance (FED):
similarly to the Fr{\'e}chet inception distance (FID) introduced in \citet{heusel2017gans}, FED is the Fr{\'e}chet distance~\citep{frechet1957distance,vaserstein1969markov} between the set of training and test embedding vectors under the Gaussian distribution assumption.
For conciseness, let $\bar{f}_{i} \,{=}\, f(x_{i}) / \lVert f(x_{i}) \rVert$ be an $\ell_{2}$ normalized embedding vector;
we normalize embedding vectors as we do when we measure the cosine similarity.
Then, with the estimated mean
$m = \frac{1}{N} \sum_{i=1}^{N} \bar{f}_{i}$
and the estimated covariance
$S = \frac{1}{N} \sum_{i=1}^{N} (\bar{f}_{i} - m) (\bar{f}_{i} - m)^{\top}$,
the FED can be defined as
\begin{align}
d^{2} \big( ( m^{\texttt{tr}}, S^{\texttt{tr}} ), ( m^{\texttt{te}}, S^{\texttt{te}} )  \big) = 
\lVert m^{\texttt{tr}} - m^{\texttt{te}} \rVert^{2}
+ \text{Tr} \big( S^{\texttt{tr}} + S^{\texttt{te}} - 2 ( S^{\texttt{tr}} S^{\texttt{te}} )^{\frac{1}{2}} \big).
\label{eq:fed}
\end{align}
As shown in Table~\ref{tb:fed}, {\imix} improves FED over contrastive learning, regardless of the distribution shift.
Note that the distance is large when the training dataset of the downstream task is the same with that of the pretext task.
This is because the model is overfit to the training dataset, such that the distance from the test dataset, which is unseen during training, has to be large.

On the other hand, Table~\ref{tb:fed} shows that {\imix}
reduces the gap between the training and test accuracy.
This implies that {\imix} is an effective regularization method for pretext tasks, such that the learned representation is more generalizable on downstream tasks.

\end{document}